\documentclass{article} 
\usepackage{preprint,times}

\usepackage{amsmath,amsfonts,bm}

\def\eqref#1{equation~\ref{#1}}

\def\1{\bm{1}}

\DeclareMathAlphabet{\mathsfit}{\encodingdefault}{\sfdefault}{m}{sl}
\SetMathAlphabet{\mathsfit}{bold}{\encodingdefault}{\sfdefault}{bx}{n}

\usepackage{booktabs}
\usepackage{graphicx}
\usepackage{tikz}
\usetikzlibrary{calc}
\usepackage{mathtools}
\usepackage{float}
\usepackage{capt-of}
\usepackage{wrapfig}
\usepackage{xcolor}
\usepackage{hyperref}
\usepackage{url}
\usepackage[normalem]{ulem}

\definecolor{viewcolor}{HTML}{448D9D}
\definecolor{sensorcolor}{HTML}{408D27}
\definecolor{editcolor}{HTML}{CF5D6F}
\definecolor{inputcolor}{HTML}{104F67}
\definecolor{initcolor}{HTML}{560D75}
\definecolor{optimcolor}{HTML}{991D40}
\definecolor{infercolor}{HTML}{285619}
\definecolor{datacolor}{HTML}{707070}
\definecolor{gradcolor}{HTML}{CC0000}
\definecolor{linkblue}{HTML}{1F6FEB}

\providecommand{\meanstd}[2]{%
  \ensuremath{#1\,{\scriptstyle\pm #2}}%
}
\newcommand{\methodname}{DyRAD\xspace}

\newcommand{\projectpage}{\href{https://dyrad-nvs.github.io}{\textcolor{linkblue}{\uline{project page}}}}
\newcommand{\projecturl}{\href{https://dyrad-nvs.github.io}{\textcolor{linkblue}{\uline{\texttt{https://dyrad-nvs.github.io}}}}}
\newcommand{\codeurl}{\href{https://github.com/Dyrad-NVS/DyRAD}{\textcolor{linkblue}{\uline{\texttt{github.com/Dyrad-NVS/DyRAD}}}}}

\providecommand{\ablationfit}[1]{%
    \leavevmode
    \begingroup
    \setbox0=\hbox{#1}%
    \ifdim\wd0>\linewidth
        \resizebox{\linewidth}{!}{\box0}%
    \else
        \box0%
    \fi
    \endgroup
}

\usepackage{xspace}

\usepackage{enumitem}

\usepackage{xcolor}
\usepackage{multirow}

\renewcommand{\paragraph}[1]{\noindent\textbf{#1}\hspace{1.0mm}}

\title{DyRAD: Radar Novel View Synthesis for Dynamic Driving Scenes}

\author{
\begin{tabular}[t]{@{}c@{}}
Merav Keidar$^{1}$ \quad
Tomer Borreda$^{1}$ \quad
Rajalakshmi Nandakumar$^{2}$ \quad
Or Litany$^{1,3}$
\\[0.6em]
\normalfont
$^{1}$Technion \qquad
$^{2}$Cornell Tech \qquad
$^{3}$NVIDIA
\end{tabular}
}

\begin{document}

\maketitle
\fancyhead{}
\renewcommand{\headrulewidth}{0pt}
\vspace{-3em}

\begin{figure}[H]
    \centering
    \includegraphics[width=\linewidth]{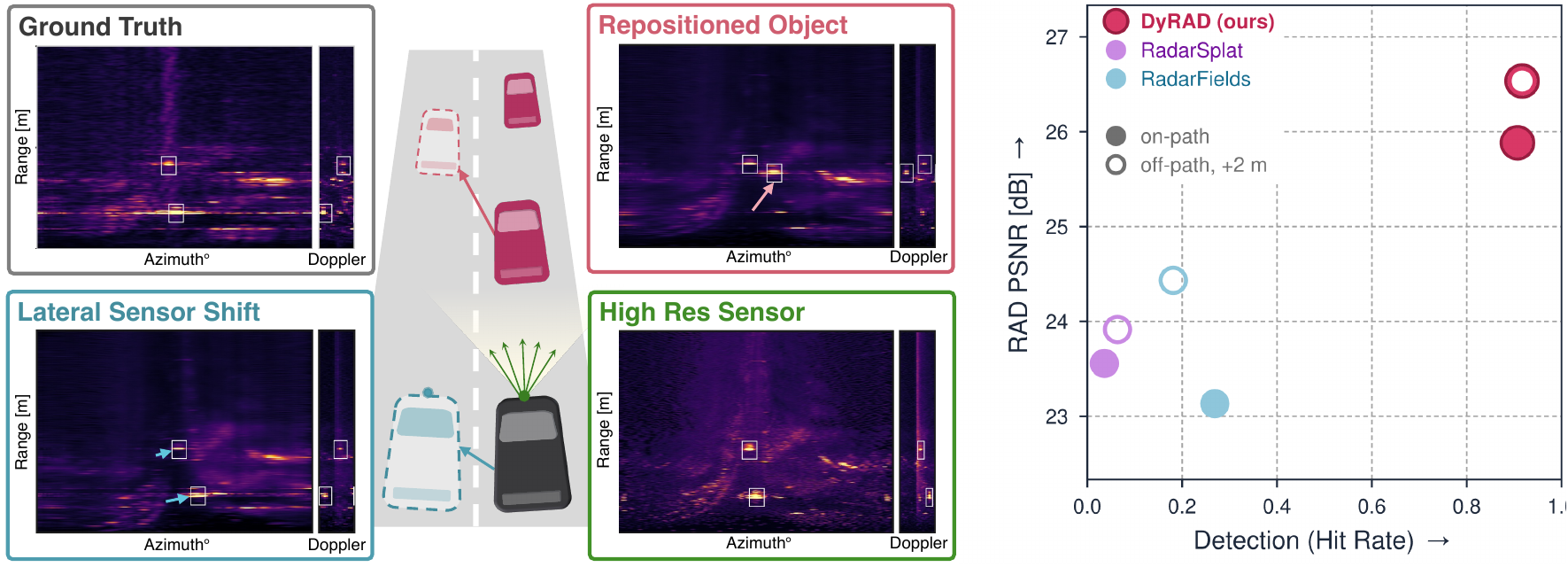}
    \caption{\textbf{Radar re-simulation with \methodname.}
    From recorded radar measurements, poses, and object boxes,
    \methodname reconstructs dynamic scenes supporting
    \textbf{\textcolor{viewcolor}{novel viewpoints}},
    \textbf{\textcolor{sensorcolor}{sensor-configuration changes}},
    and \textbf{\textcolor{editcolor}{scene editing}}.
    \textbf{Right:} Higher RAD PSNR and detection hit rate than
    both baselines on RADIal, on-path and after a
    $2\,\mathrm{m}$ lateral-shift round trip. 
    See the \projectpage{} at \projecturl{} for videos, code, and more results.}
    \label{fig:teaser}
\end{figure}

\begin{abstract}
Reconstructing dynamic driving scenes from recorded sensor data supports closed-loop evaluation of autonomous driving systems by synthesizing observations beyond the original trajectory. Unlike cameras and LiDAR, radar measures radial velocity directly through Doppler. Yet existing radar novel-view synthesis fails to exploit this capability: methods addressing dynamic scenes reconstruct only range--azimuth tensors, while methods that render Doppler assume static scenes. Moreover, because radar processing spreads each reflection across multiple bins, existing representations absorb this spread into scene geometry, causing it to render incorrectly when the viewpoint moves.
We present \methodname{}, which models dynamic driving scenes using static background reflectors and motion-tracked dynamic point reflectors to render complete range--azimuth--Doppler (RAD) tensors. Reflector velocities are derived from object tracks and projected onto the line of sight, making Doppler both a rendered output and supervision for those tracks. Crucially, we render reflectors through a fixed analytic point-spread function (PSF) derived from the radar's signal-processing chain, preventing sensor-induced spread from being baked into the scene representation.
Beyond improving scene reconstruction, this separation also enables zero-shot sensor-configuration transfer, allowing the same reconstructed scene to be rendered under different radar specifications without refitting.
We evaluate \methodname{} on RADIal, Boreas, and a synthetic benchmark across both on-path poses and displaced viewpoints untested by prior work. On RADIal, \methodname{} recovers radar detections in $90.7\%$ of reference-detected objects, compared with $26.9\%$ for the strongest baseline. 
\end{abstract}
    
\section{Introduction}

Accurate sensor simulation is critical for training and evaluating automotive systems. Reconstructing dynamic driving scenes from recorded sensor data supports this goal by enabling realistic observations to be synthesized beyond the original trajectory. 
Capturing radar measurements is particularly important given radar’s distinct capabilities within a vehicle’s sensor suite: direct radial velocity measurements via Doppler and robust operation under adverse weather. Yet its sparse measurements and limited spatial resolution make reconstruction particularly challenging. Received power varies with range and viewing aspect, while Doppler shifts depend on relative sensor--object motion. Re-simulation must therefore recover the scene from measurements that couple sensor pose, object dynamics, and signal processing.

Neural scene representations such as NeRF~\citep{nerf} and 3D Gaussian Splatting~\citep{3dgs} now support novel-view synthesis (NVS) of dynamic driving scenes for camera and LiDAR~\citep{streetgaussians,omnire,simuli}, with related approaches emerging for radar~\citep{radarfields,radarsplat,rf4d,dart}. Three capabilities, however, remain unexplored: 1) reconstructing dynamic scenes together with their Doppler signature, 2) recovering scene structure rather than the sensor’s own response, and 3) evaluating synthesis beyond the recorded trajectory.

\noindent\textbf{1) Dynamics and Doppler.} Radar measures motion
through Doppler, a property routinely exploited in radar perception~\citep{doppdrive, radarocc, radarpillars, kradar}. Yet existing radar NVS methods discard this measurement where it matters most: methods addressing dynamic driving scenes reconstruct only range--azimuth (RA), omitting Doppler~\citep{rf4d}. Methods that render Doppler assume static environments, where it arises solely from sensor motion~\citep{dart,radarsim,radarsplatrio}. No existing radar NVS method reconstructs dynamic driving scenes from complete RAD measurements while using rendered Doppler to constrain object motion.

\noindent\textbf{2) Separating the scene from the sensor.} A radar measurement is not a direct picture of the scene: the radar's signal-processing chain spreads each reflector's return across range, azimuth, and Doppler bins, as described by its point-spread function (PSF). Although existing methods estimate occupancy or transmittance to recover scene structure \citep{radarsplat, radarfields}, the spatial extent of that structure is learned directly from the measurements. A broad return can therefore be explained by a broad scene element rather than the sensor's PSF, leaving scene structure entangled with sensor-induced spread. Once absorbed into the scene representation, this spread behaves like static geometry rather than a measurement artifact, causing it to render incorrectly from displaced viewpoints. Recovering true scene structure therefore requires explicitly modeling the sensor’s response from its signal-processing chain. Furthermore, isolating the sensor response in this way unlocks a capability prior methods cannot provide: resimulating the recovered scene under a different sensor configuration by simply swapping the analytic PSF.

\noindent\textbf{3) Evaluating away from the recorded path.} Existing radar NVS evaluates only at held-out frames along the driven trajectory~\citep{radarfields,radarsplat,rf4d}. Because these test frames sit right between nearly identical training poses, interpolating directly in measurement space can score just as well as a correct forward model. As a result, this protocol fails to test synthesis at displaced viewpoints, which are precisely what closed-loop simulation requires.

To close these gaps we propose \methodname , a dynamic point-reflector scene representation coupled with a differentiable, physics-grounded RAD renderer.  Radar returns originate from
discrete scattering centers, so we represent the scene as
zero-extent point reflectors whose positions, base reflected power, and aspect-dependent reflectivity are optimized from the measurements. Static reflectors represent the background, while dynamic reflectors follow learned rigid object tracks initialized from bounding boxes.
We derive reflector velocities from these tracks and project relative sensor--reflector motion onto the line of sight to render Doppler. Jointly optimizing the reflectors and tracks against recorded RAD measurements makes Doppler both a rendered output and a constraint on those tracks.

We address the second gap by explicitly modeling the sensor's response.
Each point reflector is rendered through a sensor-specific PSF across range, azimuth, and Doppler, derived from the radar's signal-processing chain and held fixed throughout optimization. 
Using zero-extent reflectors with a fixed sensor response constrains how measurement spread is explained, reducing ambiguity between scene structure and sensor-induced blur. This makes rendering at sensor poses away from the driven trajectory a forward model rather than interpolation in measurement space, and it makes the recovered scene portable across sensors (Fig.~\ref{fig:teaser}).

We evaluate DyRAD on RADIal, Boreas, and a synthetic benchmark,
testing reconstruction at held-out poses along the
recorded trajectory as well as at displaced viewpoints; the latter addressing the third gap. We assess
displaced views directly against synthetic ground truth and through a render-and-refit cycle on real data.
On RADIal, \methodname{} increases full-RAD correlation from $0.068$ to $0.272$ and foreground hit rate from $26.9\%$ to $90.7\%$ over the strongest respective baselines.
Ablations show that Doppler supervision reduces vehicle Doppler
peak error by $60\%$ relative to RA-only fitting, while fixing
the analytic PSF more than doubles joint detection recall
compared with learning Gaussian extents or the PSF.
Finally, the same reconstructed scene supports coarse-to-fine
sensor-configuration transfer without refitting, improving
object-region RAD PSNR and nearly doubling detection F1
over linear upsampling.

Our main contributions are as follows:

\begin{enumerate}[leftmargin=*, noitemsep, topsep=0pt, partopsep=0pt] %
\item \textbf{Physics-grounded RAD rendering of dynamic scenes.}
To our knowledge, we introduce the first radar NVS method
to render Doppler for dynamic driving scenes.
We reconstruct scenes as static and dynamic point reflectors following rigid object tracks and render complete RAD tensors with Doppler derived from relative sensor--reflector motion.

\item \textbf{Decoupling scene structure from sensor response.}
We render point reflectors through a fixed, sensor-specific PSF,
separating scene structure from sensor-induced spread. This improves object-region reconstruction and detection
while also enabling the same scene to be rendered under alternative
radar configurations without refitting.

\item \textbf{Off-path evaluation for radar novel-view synthesis.}
Prior radar NVS evaluates only along the recorded trajectory, testing interpolation rather than true spatial generalization. We introduce off-path evaluation via displaced ground-truth views in a synthetic benchmark  and a cycle-consistency protocol adapted to real radar recordings.
\end{enumerate}

\section{Related Work}
\label{sec:related}

\subsection{Novel-view synthesis for dynamic driving scenes}
Reconstructing driving scenes from recorded data has become an established approach to sensor simulation for autonomous driving. Once reconstructed, a scene can be rendered along new ego trajectories, enabling training and closed-loop evaluation beyond the recorded drive~\citep{blocknerf,lidarsim,unisim,neurad}. Early neural approaches focused on static camera scenes~\citep{blocknerf,urf}, later extending to dynamic driving environments~\citep{suds,emernerf,streetgaussians,omnire}. Reconstruction-based simulation also expanded to LiDAR, progressing from static~\citep{lidarsim,nfl} to dynamic scenes~\citep{dynfl,lidar4d}, and subsequently to joint camera--LiDAR rendering~\citep{unisim,neurad,splatad,simuli}. For camera and LiDAR, dynamic driving scenes can now be reconstructed and replayed from viewpoints that were never driven.
Our work makes this possible also for radar.

\subsection{Radar reconstruction and Doppler}
\label{sec:related-radar}

\paragraph{Radar novel-view synthesis.}
Radar reconstruction has developed along similar lines as its camera and LiDAR counterparts. Radar Fields~\citep{radarfields} and RadarSplat~\citep{radarsplat} reconstruct RA measurements of \emph{static} driving scenes, the former with a frequency-space neural field, the latter with Gaussian primitives combining modeling
of radar noise and multipath effects. RF4D~\citep{rf4d} extends the Radar Fields framework to dynamic
driving scenes through a spatiotemporal field and a scene-flow module that predicts motion offsets between adjacent frames.
Yet it too reconstructs RA alone.
NeuRadar~\citep{neuradar} extends joint camera--LiDAR rendering of dynamic driving scenes to radar point clouds, but renders radar as sparse detections without Doppler information rather than as raw, dense measurements.

\paragraph{Doppler as a motion cue.}
Doppler has been used as an input motion cue for visual scene reconstruction, as in 4DRadar-GS~\citep{4dradargs}, which uses per-point radar velocities but does not render radar measurements. Other methods render Doppler but assume static environments~\citep{dart,radarsim,radarsplatrio}, where it arises solely from sensor motion. We instead reconstruct dynamic driving scenes directly from full RAD measurements, where Doppler is both rendered and used to refine the object motion that produced it.

\subsection{Sensor modeling in neural rendering}
\label{sec:related-sensor}

Neural rendering has repeatedly improved by replacing an idealized sensor with a model of how the measurement is actually formed. Methods account for sampling footprints and beam divergence~\citep{mipnerf,nfl}, processing and blur~\citep{rawnerf,deblurnerf}, and appearance variation~\citep{nerfw}.
Radar reconstruction additionally accounts for propagation, antenna gain, and range--Doppler geometry ~\citep{radarfields,dart,radarsim,mmir}, as well as speckle and multipath~\citep{radarsplat}.
Measurement spreading has been addressed through PSF convolution for simulation~\citep{radsimreal}, range smoothing and antenna-profile convolution~\citep{radarsplat}, and Doppler soft binning~\citep{radarsplatrio}.
We instead represent the scene using static and dynamic point reflectors coupled with a fixed, sensor-specific RAD PSF, explicitly separating sensor-induced measurement spread from the learned scene representation.

\section{Method}
\label{sec:method}

\begin{figure}[t]
    \centering
    \includegraphics[width=\columnwidth]{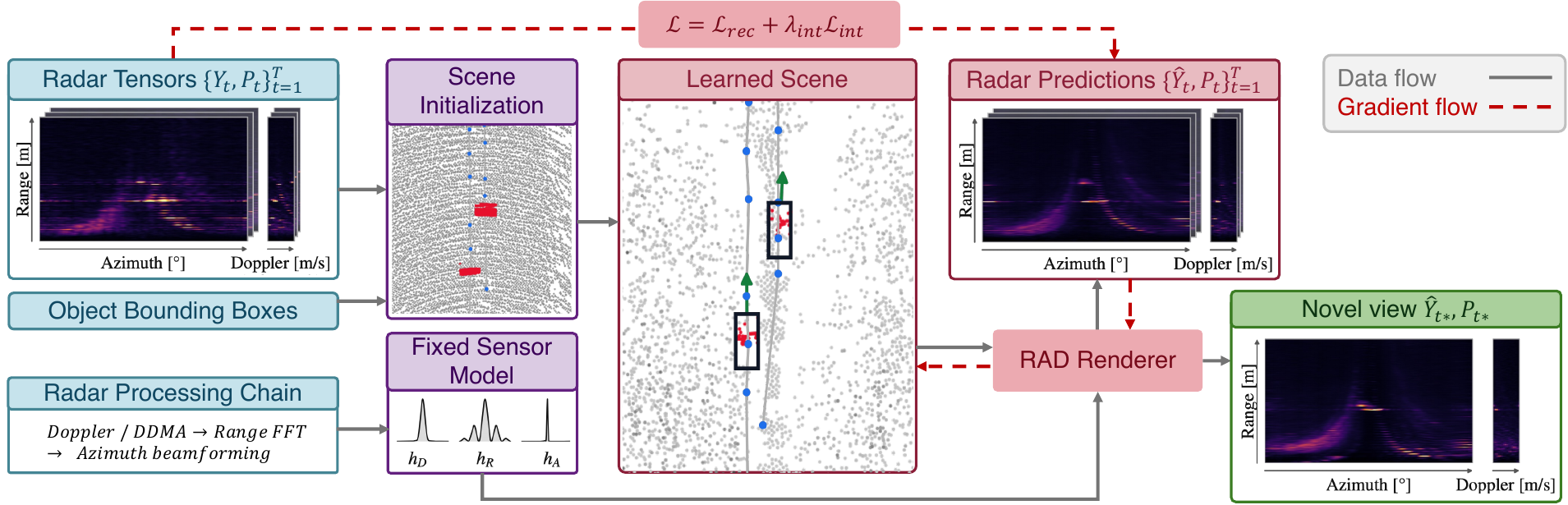}
    \caption{
    \textbf{\methodname\ overview.} \textbf{\textcolor{inputcolor}{Inputs:}} recorded RAD tensors, sensor poses, object bounding-boxes, and radar processing information when available.
    \textbf{\textcolor{initcolor}{Initialization:}} measurements and annotations initialize the scene; processing information defines the fixed sensor response.
    \textbf{\textcolor{optimcolor}{Optimization:}} the differentiable RAD renderer jointly fits reflectors and object tracks to recorded measurements.
    \textbf{\textcolor{infercolor}{Inference:}} the optimized scene produces RAD predictions $\hat{\mathbf{Y}}(t^\star,\mathbf{P}^\star)$ at query times and sensor poses.
    Solid gray arrows show \textbf{\textcolor{datacolor}{data flow}}; dashed red arrows show \textbf{\textcolor{gradcolor}{gradient flow}}.
    }
    \label{fig:method-overview}

\end{figure}

\paragraph{Method Overview.}
\methodname{} reconstructs dynamic driving scenes from radar measurements to synthesize RAD tensors at novel sensor poses (Fig.~\ref{fig:method-overview}). We represent the scene using static and dynamic point reflectors, with dynamic reflectors following rigid object tracks (Sec.~\ref{sec:scene-representation}). We couple this representation with a differentiable renderer that maps reflectors to radar measurements through a fixed sensor-specific response (Sec.~\ref{sec:rad-rendering}). We initialize the reflectors and tracks from radar measurements and object bounding boxes, 
then jointly optimize them to reproduce the observations (Sec.~\ref{sec:scene-Initialization}).

\paragraph{Problem Setting.}
We are given a sequence of radar measurements $\{\mathbf{Y}_t\}_{t=1}^{T}$ and corresponding sensor poses $\{\mathbf{P}_t\}_{t=1}^{T}$. 
Each RAD tensor $\mathbf{Y}_t\in\mathbb{R}_{\geq 0}^{D\times R\times A}$ records return strength over $D$ radial-velocity bins, $R$ range bins, and $A$ azimuth bins.
We obtain RA and RD projections by averaging over Doppler and azimuth, respectively; for sensors without Doppler, we reconstruct RA maps.
We additionally assume dynamic-object bounding-box annotations at a subset of frames, a common input in dynamic driving-scene reconstruction~\citep{streetgaussians,omnire}.
These annotations initialize object tracks, providing spatial and motion information that is difficult to infer from sparse radar returns alone. 
Since the measurements do not resolve elevation, we model geometry and motion in the ground plane, with sensor poses defined by planar position and heading.
Our goal is to synthesize measurements $\hat{\mathbf{Y}}(t^\star,\mathbf{P}^\star)$ at a query time $t^\star$ and sensor pose $\mathbf{P}^\star$.

\subsection{Scene Representation }
\label{sec:scene-representation}

\paragraph{Scene Primitives.}
We represent the scene as $N$ point reflectors with learned positions and reflectivities. Reflectors have no spatial extent; the renderer models sensor-induced spreading through a fixed PSF.
We partition the reflectors into a static background and $M$ dynamic objects, with reflectors within each object sharing a rigid motion. 
The scene is then: 
$\mathcal{S}
    = \left\{
        \bigl(\mathbf{x}_i(\cdot), o_i,
        \boldsymbol{\eta}_i, q_i\bigr)
    \right\}_{i=1}^{N}, $
where $\mathbf{x}_i(t)\in\mathbb{R}^{2}$ is reflector $i$'s world-space position at time $t$. The learned logit $o_i\in\mathbb{R}$ defines reflector $i$'s base reflected power $\alpha_i=\sigma(o_i)\in(0,1)$. $\boldsymbol{\eta}_i\in\mathbb{R}^{K}$ contains spherical-harmonic coefficients for view-dependent reflectivity. The assignment $q_i\in\{0,\ldots,M\}$ identifies the static background ($q_i=0$) or a dynamic object ($q_i>0$).

\paragraph{Motion Modeling.}
Following prior dynamic driving-scene reconstruction~\citep{streetgaussians,omnire}, we represent each moving object by a planar rigid track (Fig.~\ref{fig:motion-model}).
\par\vspace{-0.4em}\noindent
\begin{minipage}[t]{0.50\linewidth}
\vspace{0pt}
Object $j$ has one learned track point $\mathbf{c}_{j,k}\in\mathbb{R}^{2}$ per training timestamp $t_k$, with the first point defining its reference-frame origin. For $t\in[t_{k},t_{k+1}]$, we linearly interpolate $\mathbf{c}_{j,k}$ and $\mathbf{c}_{j,k+1}$ to obtain its position $\mathbf{p}_j(t)$. We estimate its orientation from the direction of travel using neighboring track points and interpolate it between timestamps. The rotation $\mathbf{R}_j(t)$ describes the change in orientation relative to the object's initial orientation.
Reflector $i$'s world-space position is then:
\begin{equation}
    \mathbf{x}_i(t)=
    \begin{cases}
        \mathbf{x}_i,
        & q_i=0, \\[2mm]
        \mathbf{p}_j(t)+\mathbf{R}_j(t)\boldsymbol{\mu}_i,
        & q_i=j>0.
    \end{cases}
    \label{eq:rigid-track}
\end{equation}
where $\mathbf{x}_i$ is a static reflector's learned world-space position and $\boldsymbol{\mu}_i$ is a dynamic reflector's learned position in its object's reference frame.

\end{minipage}\hfill
\begin{minipage}[t]{0.48\linewidth}
\vspace{0pt}
\centering
\includegraphics[width=\linewidth]
    {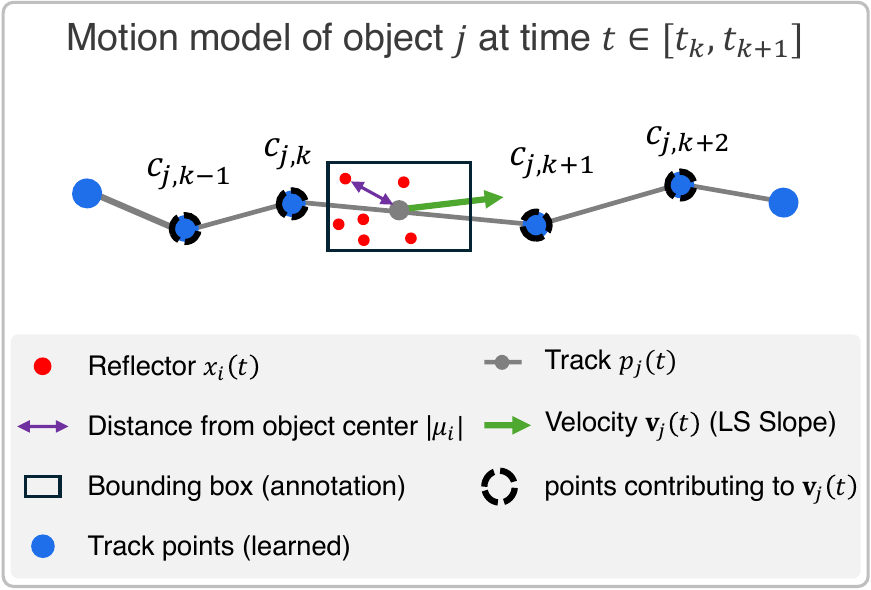}
\captionof{figure}{\textbf{Object motion along a learned track.}  Four neighboring track points estimate translational velocity. Reflectors follow the track's translation and relative rotation, preserving their object-frame coordinates $\boldsymbol{\mu}_i$.}
\label{fig:motion-model}
\end{minipage}
\par\vspace{-0.4em}\noindent
\noindent

\subsection{Physics-Grounded RAD Rendering}
\label{sec:rad-rendering}
We render RAD tensors by projecting reflectors into sensor coordinates, computing their Doppler, applying the sensor-specific PSF, and accumulating their contributions. The pipeline is differentiable and implemented in CUDA for efficiency.

\paragraph{Reflector Projection.}
At time $t$, we transform each reflector's world-space position into the sensor frame using pose $\mathbf{P}$ and compute its range $r_i$, azimuth $\phi_i$, and unit viewing direction $\mathbf{d}i$ from the sensor to the reflector.
We model view-dependent reflected power as:
$
    \rho_i(\mathbf{d}_i)
    =
    \alpha_i\,
    \exp\!\left(
        \sum_{k=0}^{K-1}
        \eta_{i,k}B_k(\mathbf{d}_i)
    \right),
    \label{eq:reflector-power}
$
where $B_k$ is the $k$-th real spherical-harmonic (SH) basis function. We fix the degree-zero coefficient to $\eta_{i,0}=0$ to avoid redundancy between the constant harmonic and the base reflected power $\alpha_i$.

\paragraph{Doppler.}
Doppler measures relative sensor--reflector velocity along the line of sight. 
Let $\mathbf{v}_i(t)$ denote reflector $i$'s world-space velocity. Static reflectors have $\mathbf{v}_i(t)=\mathbf{0}$. For dynamic reflectors, we derive $\mathbf{v}_i(t)$ from the learned object track, accounting for changes in both position and orientation. We estimate the object's translational velocity as the slope of a linear fit to four neighboring object track points (Fig.~\ref{fig:motion-model}) and additionally include the velocity induced by changes in object orientation.
We obtain ego sensor velocity $\mathbf{v}_s(t)$ from the dataset when available and otherwise estimate it from timestamped sensor poses. A reflector's radial velocity is then  $\nu_i(t)=s_D\cdot\hat{\mathbf{r}}_i(t)^\top\bigl(\mathbf{v}_i(t)-\mathbf{v}_s(t)\bigr)$, where $\hat{\mathbf{r}}_i(t)$ is the unit vector from the sensor to reflector $i$ and $s_D\in\{-1,1\}$ accounts for the sensor's Doppler sign convention.

Radar Doppler measurements are periodic: radial velocities outside the sensor's unambiguous interval wrap back into it, so different radial velocities can produce the same Doppler coordinate. We reproduce this behavior by wrapping $\nu_i(t)$ into this interval, yielding the rendered coordinate $\widetilde{\nu}_i(t)$.

\paragraph{Sensor Response.}
The sensor's PSF determines how each reflector's return spreads across range, azimuth, and Doppler axes. We model reflector $i$'s contribution as:
\begin{equation}
    \psi_i(r,\phi,\nu)
    =
    \rho_i(\mathbf{d}_i)\,
    h_R(r-r_i)\,
    h_A(\phi,\phi_i)\,
    h_D(\nu-\widetilde{\nu}_i(t)),
    \label{eq:reflector-response}
\end{equation}
where $\rho_i(\mathbf{d}_i)$ is its view-dependent reflected power, and $h_R$, $h_A$, and $h_D$ describe the sensor's spread kernel along each axis, where each captures range-FFT leakage, the azimuth beamformer response, and the Doppler-processing response, respectively. 
The Doppler kernel $h_D$ is periodic, wrapping responses across the measured interval's boundaries.
The azimuth kernel depends on both the evaluated angle $\phi$ and the reflector angle $\phi_i$, allowing the PSF to vary across viewing directions.

All kernels are computed before scene optimization and remain fixed. We derive them from the signal-processing chain when available, and approximate otherwise (Appendix Sec.~\ref{sec:supp-psf}).

\paragraph{Rasterization.}
Unlike depth-ordered alpha compositing in camera rendering, radar signal formation involves the superposition of echoes from multiple reflectors. We approximate it
by summing reflected powers incoherently.
We form the RAD tensor by summing the contributions of all
reflectors' responses
at the sensor's sampling grid:
$
    \boldsymbol{\Psi}(t,\mathbf{P})[\ell,m,n]
    =
    \sum_{i=1}^{N}
    \psi_i(r_m,\phi_n,\nu_\ell),
    \label{eq:rad-rasterization}
$
where $r_m$, $\phi_n$, and $\nu_\ell$ are the range, azimuth, and Doppler coordinates of cell $[\ell,m,n]$.

Finally, we convert this power tensor to the recorded intensity representation (e.g., amplitude, power, or log-compressed power), incorporating a sensor-specific measurement floor to account for the nonzero background. This floor is determined during preprocessing and held fixed throughout optimization. The resulting tensor is our prediction $\hat{\mathbf{Y}}(t,\mathbf{P})$.

\subsection{Scene Initialization and Optimization}
\label{sec:scene-Initialization}
\label{sec:scene-optimization}

\paragraph{Initialization.}
We associate object bounding boxes across frames to initialize tracks and use the boxes to separate dynamic-object regions from the static background.
Each track contains one point per training frame, initialized from annotated locations where available and linearly interpolated otherwise.
We initialize background reflectors from peaks in Doppler-averaged RA maps.
For each dynamic object, we extract peaks within its bounding box from the Doppler bin with the strongest response at its annotated location.

We project the peaks onto the ground plane and accumulate them across frames retaining background peaks in world coordinates and aligning dynamic peaks to their object's reference frame using the initialized tracks. We merge nearby peaks with the same object assignment to obtain the initial reflectors. 
Appendix Sec.~\ref{sec:supp-scene-initialization} provides details and an illustration. 

\paragraph{Optimization.}
We jointly optimize the reflector parameters and track points using
$
    \mathcal{L}
    =
    \mathcal{L}_{\mathrm{rec}}
    + \lambda_{\mathrm{int}}\mathcal{L}_{\mathrm{int}},
$
where $\mathcal{L}_{\mathrm{rec}}$ is the mean squared error between predicted and recorded RAD tensors (or RA maps for sensors without Doppler). Inspired by RegNeRF~\citep{regnerf}, the interpolation-consistency loss $\mathcal{L}_{\mathrm{int}}$ regularizes predictions between training frames. 
We sample a time $\tilde{t}$ between adjacent training timestamps $t_0$ and $t_1$ and interpolate the sensor pose to obtain $\tilde{\mathbf{P}}$.
We compare the predicted RA map with the neighboring measurements' average after spatial smoothing:
\begin{equation}
    \mathcal{L}_{\mathrm{int}}
    =
    \left\|
        \mathcal{G}\!\left(
            \Pi_{\mathrm{RA}}\bigl(
                \hat{\mathbf{Y}}(\tilde{t},\tilde{\mathbf{P}})
            \bigr)
        \right)
        -
        \mathcal{G}\!\left(
            \frac{
                \Pi_{\mathrm{RA}}(\mathbf{Y}_{t_0})
                + \Pi_{\mathrm{RA}}(\mathbf{Y}_{t_1})
            }{2}
        \right)
    \right\|_1,
    \label{eq:interpolation-loss}
\end{equation}
where $\Pi_{\mathrm{RA}}$ averages over Doppler and $\mathcal{G}$ applies a fixed spatial Gaussian blur.
This provides a coarse appearance constraint at intermediate timesteps. We apply it only to RA, since the neighboring-frame average does not specify intermediate object positions or Doppler profiles. We evaluate its effect in the ablation studies (Sec. \ref{sec:ablations}), with full results in Appendix Sec.~\ref{sec:suppl-ablations}. 

\paragraph{Densification.}
During optimization, we periodically extract peaks from positive reconstruction residuals and add static-background reflectors where no nearby reflector exists. 
Additional details  and hyperparameters are provided in Sec.~\ref{sec:supp-densification} of the appendix.

\section{Experiments}
\label{sec:experiments}
We encourage readers to explore the \projectpage, which provides videos of all real-data off-path and synthetic experiments.
\subsection{Experimental Setup}
\label{sec:experimental-setup}

\paragraph{Datasets.} We evaluate on two real-world datasets, \emph{RADIal}~\citep{radial}
and \emph{Boreas}~\citep{boreas}, and one synthetic benchmark. Our primary benchmark uses ten 60-frame RADIal sequences, with raw radar measurements, vehicle annotations, and ego-motion measurements, supporting full-RAD reconstruction and evaluation of scene dynamics.
For Boreas, we use three 50-frame sequences. Its $360^\circ$ spinning Navtech radar provides RA measurements without Doppler, testing reconstruction in the baselines' native sensor setting and complementing RADIal's front-facing radar. Finally, we construct five synthetic scenes with an idealized RADIal-format sensor model to obtain ground-truth measurements at displaced poses for direct off-path evaluation. We will release the synthetic benchmark, including off-path ground-truth
measurements, to support future research. Implementation details and the generation procedure are provided in Appendix Sec.~\ref{sec:supp-synthetic}.

\paragraph{Baselines.}
We compare with RadarSplat~\citep{radarsplat}, and RadarFields~\citep{radarfields}, which reconstruct static scenes and predict RA measurements. We use their published configurations on Boreas and adapt them to RADIal's sensor geometry and resolution.
For RAD reconstruction and detection evaluation, we assign Doppler assuming a static scene and using the sensor's ego-motion; these results are marked with an asterisk.
We omit RF4D \citep{rf4d} because its released code is “still under checking”.
We also evaluate \methodname-static, which retains our point-reflector representation and renderer but treats all reflectors as static and omits bounding-box separation during initialization. Comparing this variant with the full model measures the combined benefit of object separation and motion modeling.

\paragraph{Metrics.}
\emph{Reconstruction.} Following RadarSplat and RadarFields, we report PSNR and SSIM~\citep{ssim} over full measurements and object regions of RAD, RA, and RD, and full-measurement LPIPS~\citep{lpips} over full RA and RAD measurements.
However, agreement at the noise floor can dominate these global scores in sparse radar data (Sec.~\ref{sec:suppl-control}).
We therefore report Pearson correlation, a scale-invariant measure of structural agreement. $\rho$ and $\rho_{\mathrm{obj}}$ denote correlation over full measurements and object regions, respectively.

\emph{Detection.} To assess whether synthesized measurements preserve usable radar returns, we apply identical RD-CFAR detection to reference and synthesized
RAD tensors and compare the resulting point clouds using metrics adapted from RadarGen~\citep{radargen}.
We report spatial Chamfer distance (CD), joint F1 over position, Doppler, and power, and foreground hit rate: the fraction of reference-occupied annotated boxes also containing predicted detections.
Detection evaluation covers RADIal and synthetic data; Boreas lacks the Doppler dimension required by RD-CFAR. Metric definitions and additional results appear in Appendix Sec. ~\ref{sec:supp-experiments}.

\subsection{Novel View Synthesis}
\label{sec:nvs}
\begin{figure*}[t]
    \centering
    \includegraphics[width=\columnwidth]{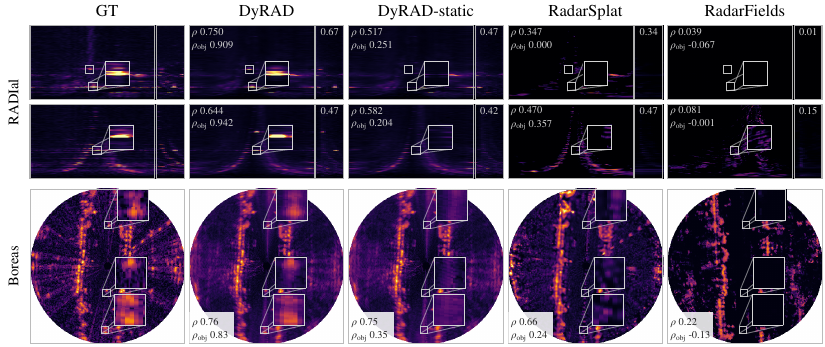}
    \caption{\textbf{On-path reconstruction on RADIal and Boreas.}
    The top two rows show held-out RADIal frames with RA (left) and RD (right) projections; the bottom row shows a held-out Boreas RA frame in Cartesian bird's-eye view.
    White boxes mark objects, with selected regions enlarged in the insets.
    \methodname\ preserves object returns that the static variant
    and baselines weaken or miss, even when surrounding scene structure appears similar.}
    \label{fig:onpath_qualitative}
    \vspace{-4pt}
\end{figure*}

\paragraph{On-Path Evaluation.}
Following prior work~\citep{radarsplat,radarfields}, we hold out every fifth frame and exclude its measurements and bounding boxes from initialization and optimization (Table~\ref{tab:onpath_reconstruction}).
Our static variant already improves reconstruction over both baselines, including in their native RA output: on RADIal, \methodname-static achieves correlation $0.533$, compared with $0.291$ for RadarSplat and $0.055$ for RadarFields.
However, this stronger aggregate reconstruction still misses most vehicle returns.
Moving from \methodname-static to the full model increases full-RAD correlation from $0.213$ to $0.272$, while object-region RAD correlation rises from $-0.013$ to $0.658$ and foreground hit rate from $7.0\%$ to $90.7\%$.
On Boreas, the gap over RadarSplat is smaller in full-measurement correlation ($0.752$ versus $0.666$) than within object regions ($0.622$ versus $0.178$). These results support the benefit of explicit object modeling even without Doppler supervision.
Figure~\ref{fig:onpath_qualitative} provides qualitative examples on both datasets, with enlarged regions highlighting vehicle returns that \methodname\ preserves but the static variant and baselines weaken or miss.

\begin{table*}[t]
    \centering
    \setlength{\tabcolsep}{2pt}
    \renewcommand{\arraystretch}{1.0}
    \caption{\textbf{On-path evaluation.}
    Mean $\pm$ sample standard deviation across ten RADIal sequences and three Boreas sequences.
    $^{*}$ indicates Doppler lifted RA-only baselines.
    Full results in Sec.~\ref{sec:supp-onpath}.}
    \label{tab:onpath_reconstruction}
    \label{tab:onpath_radial_detection}
    \vspace{4pt}
    \ablationfit{%
    \begin{tabular}{@{}lccc@{\hspace{6pt}}ccc@{\hspace{6pt}}ccc@{}}
    \toprule
    & \multicolumn{3}{c}{RADIal: RAD reconstruction}
    & \multicolumn{3}{c}{RADIal: detection}
    & \multicolumn{3}{c}{Boreas: RA reconstruction} \\
    \cmidrule(lr){2-4}\cmidrule(lr){5-7}\cmidrule(lr){8-10}
    Method
    & PSNR$\uparrow$ & $\rho\uparrow$ & $\rho_\text{obj}\uparrow$
    & CD [m] $\downarrow$ & F1 $\uparrow$ & Hit rate $\uparrow$
    & PSNR$\uparrow$ & $\rho\uparrow$ & $\rho_\text{obj}\uparrow$ \\
    \midrule
    RadarSplat$^{*}$
    & \meanstd{23.56}{0.85}
    & \meanstd{0.068}{0.028}
    & \meanstd{-0.010}{0.057}
    & \meanstd{6.53}{5.82}
    & \meanstd{0.073}{0.048}
    & \meanstd{0.036}{0.067}
    & \meanstd{23.92}{1.07}
    & \meanstd{0.666}{0.036}
    & \meanstd{0.178}{0.249} \\
    RadarFields$^{*}$
    & \meanstd{23.13}{1.23}
    & \meanstd{0.015}{0.009}
    & \meanstd{-0.011}{0.051}
    & \meanstd{6.97}{0.96}
    & \meanstd{0.016}{0.007}
    & \meanstd{0.269}{0.245}
    & \meanstd{19.96}{0.84}
    & \meanstd{0.279}{0.078}
    & \meanstd{0.016}{0.128} \\
    \methodname-static
    & \meanstd{25.67}{0.95}
    & \meanstd{0.213}{0.054}
    & \meanstd{-0.013}{0.072}
    & \meanstd{4.71}{1.29}
    & \meanstd{0.123}{0.041}
    & \meanstd{0.070}{0.083}
    & \meanstd{25.31}{0.90}
    & \meanstd{0.749}{0.018}
    & \meanstd{0.278}{0.217} \\
    \methodname
    & \meanstd{\mathbf{25.88}}{0.95}
    & \meanstd{\mathbf{0.272}}{0.062}
    & \meanstd{\mathbf{0.658}}{0.070}
    & \meanstd{\mathbf{4.01}}{1.02}
    & \meanstd{\mathbf{0.179}}{0.041}
    & \meanstd{\mathbf{0.907}}{0.086}
    & \meanstd{\mathbf{25.36}}{0.90}
    & \meanstd{\mathbf{0.752}}{0.017}
    & \meanstd{\mathbf{0.622}}{0.130} \\
    \bottomrule
    \end{tabular}%
    }
\end{table*}

\paragraph{Real-Data Off-Path Evaluation.}
Without ground-truth measurements at displaced viewpoints, we adopt the cycle-consistency protocol of Neural LiDAR Fields~\citep{nfl}.
We fit each method to the recordings, render a trajectory shifted laterally by $2\mathrm{m}$, and train a new instance on those renders. Its predictions at the original poses and training timestamps are then compared with the real recordings; scores are therefore not directly comparable to the held-out on-path results.
This tests whether scene information remains recoverable through displaced rendering and refitting; direct off-path accuracy is evaluated separately on synthetic data.
\methodname{} achieves the best means across the metrics in Table~\ref{tab:offpath_reconstruction}.
On RADIal, it substantially outperforms the prior baselines in both reconstruction and detection, recovering detections in $91.7\%$ of reference-detected objects, compared with $18.1\%$ for the strongest prior baseline on this measure, and $20.9\%$ for the static model. 
On Boreas, RadarSplat nearly matches our PSNR ($24.29$ versus $24.51$\,dB), yet its object-region RA correlation is substantially lower ($0.211$ versus $0.641$).
The object-reconstruction benefit observed on-path thus persists through displaced rendering and refitting, including in the baselines' native RA setting without Doppler supervision.

\begin{table*}[t]
    \centering
    \footnotesize
    \setlength{\tabcolsep}{2pt}
    \renewcommand{\arraystretch}{1.0}
    \caption{\textbf{Real-data off-path evaluation.}
    Reconstruction at the original poses after a $2\mathrm{m}$
    lateral-shift round trip.
    Mean $\pm$ sample standard deviation across ten RADIal sequences and three Boreas sequences. 
    $^{*}$ indicates Doppler lifted RA-only baselines. See the \projectpage{} for off-path videos and Sec.~\ref{sec:supp-offpath} for full quantitative results.}
    \label{tab:offpath_reconstruction}
    \vspace{4pt}
    \ablationfit{%
    \begin{tabular}{@{}lccc@{\hspace{6pt}}ccc@{\hspace{6pt}}ccc@{}}
    \toprule
    & \multicolumn{3}{c}{RADIal: RAD reconstruction}
    & \multicolumn{3}{c}{RADIal: detection}
    & \multicolumn{3}{c}{Boreas: RA reconstruction} \\
    \cmidrule(lr){2-4}\cmidrule(lr){5-7}\cmidrule(lr){8-10}
    Method
    & PSNR$\uparrow$ & $\rho\uparrow$ & $\rho_\text{obj}\uparrow$
    & CD [m] $\downarrow$ & F1 $\uparrow$ & Hit rate $\uparrow$
    & PSNR$\uparrow$ & $\rho\uparrow$ & $\rho_\text{obj}\uparrow$ \\
    \midrule
    RadarSplat$^{*}$
    & \meanstd{23.92}{0.85}
    & \meanstd{0.076}{0.032}
    & \meanstd{-0.004}{0.053}
    & \meanstd{6.35}{4.75}
    & \meanstd{0.077}{0.050}
    & \meanstd{0.064}{0.065}
    & \meanstd{24.29}{1.16}
    & \meanstd{0.689}{0.042}
    & \meanstd{0.211}{0.237} \\
    RadarFields$^{*}$
    & \meanstd{24.43}{0.75}
    & \meanstd{0.023}{0.018}
    & \meanstd{-0.011}{0.048}
    & \meanstd{9.19}{6.07}
    & \meanstd{0.027}{0.020}
    & \meanstd{0.181}{0.183}
    & \meanstd{21.09}{1.28}
    & \meanstd{0.538}{0.081}
    & \meanstd{0.095}{0.123} \\
    \methodname-static
    & \meanstd{26.41}{0.73}
    & \meanstd{0.335}{0.054}
    & \meanstd{0.056}{0.064}
    & \meanstd{3.47}{0.61}
    & \meanstd{0.210}{0.042}
    & \meanstd{0.209}{0.144}
    & \meanstd{24.45}{0.62}
    & \meanstd{0.741}{0.007}
    & \meanstd{0.261}{0.215} \\
    \methodname
    & \meanstd{\mathbf{26.53}}{0.74}
    & \meanstd{\mathbf{0.363}}{0.052}
    & \meanstd{\mathbf{0.536}}{0.077}
    & \meanstd{\mathbf{3.25}}{0.57}
    & \meanstd{\mathbf{0.236}}{0.040}
    & \meanstd{\mathbf{0.917}}{0.101}
    & \meanstd{\mathbf{24.51}}{0.62}
    & \meanstd{\mathbf{0.744}}{0.009}
    & \meanstd{\mathbf{0.641}}{0.165} \\
    \bottomrule
    \end{tabular}%
    }
\end{table*}

\paragraph{Synthetic Off-Path Evaluation.}We directly assess off-path accuracy  by fitting each method to five synthetic scenes and evaluating without refitting at lateral offsets of $-1.75$, $+1.75$, and $+3.5\mathrm{m}$ and yaw offsets of $+5^\circ$ and $+10^\circ$.
Across the 25 scene--offset pairs, \methodname\ achieves the highest mean correlation scores, globally and within objects.
Object-region RAD correlation reaches $0.356$, compared with $0.161$ for \methodname-static and $0.086$ for RadarFields, the strongest prior baseline, demonstrating improved object reconstruction at displaced viewpoints.
CD is nearly identical to RadarSplat's ($5.13$ versus $5.14\mathrm{m}$), but joint detection F1 is substantially higher ($0.227$ versus $0.106$), indicating better agreement in location, Doppler, and power despite similar spatial distances.
F1 is close to \methodname-static ($0.223$), whereas foreground hit rate is much higher: $74.6\%$, compared with $22.4\%$ for \methodname-static and $36.0\%$ for RadarSplat.
However, this greater coverage comes with more reference-empty annotated boxes containing predicted detections: $4.9$ on average, versus $0.7$ and $3.4$, respectively.
Videos of all synthetic scenes and evaluated offsets appear on the \projectpage; the generator and full results are described in  Sec.~\ref{sec:supp-synthetic}.

\subsection{Additional Capabilities}
\label{sec:additional-capabilities}
Separating the scene from the sensor response enables
sensor-configuration transfer without retraining.
We demonstrate coarse-to-fine transfer using two configurations processed from the same raw ADC recordings.
After fitting to coarse measurements, we replace the PSF and sampling grid with the finer configuration's, keeping scene parameters fixed.
Against real fine-resolution measurements, direct rendering reduces CD from $7.27$ to $3.20\mathrm{m}$ and increases F1 from $0.101$ to $0.195$ over linear upsampling of the same model's coarse renders.
It also improves object-region RAD PSNR, although upsampling scores higher on most correlation and SSIM metrics.
These detection gains without refitting suggest that the learned reflectors capture scene structure beyond the coarse measurement resolution.
Details and full results appear in Appendix~\ref{sec:supp-config-transfer}.
Additionally, the explicit scene and motion representation supports scene editing, demonstrated qualitatively in Fig.~\ref{fig:teaser}.

\subsection{Ablation Studies}
\label{sec:ablations}
We ablate Doppler supervision, interpolation consistency, and PSF modeling under the RADIal on-path and off-path protocols.
Complete results appear in Tables ~\ref{tab:suppl_abl_onpath_reconstruction}~\ref{tab:suppl_abl_offpath_detection} in Sec.~\ref{sec:suppl-ablations} of the appendix.

\paragraph{Doppler supervision.}
Doppler supervision raises joint detection F1 from $0.144$ to $0.179$ on-path and from $0.120$ to $0.236$ off-path, while reducing object Doppler peak error from $1.843$ to $0.732$ bins and from $2.264$ to $1.375$ bins, respectively. Yet foreground hit rate remains nearly unchanged under both protocols: bounding-box initialization and RA-only fitting already recover vehicle coverage, while Doppler supervision refines the velocity structure of those returns. Object-region RAD correlation also increases from $0.504$ to $0.658$ on-path and from $0.389$ to $0.536$ off-path, at the cost of reductions of $0.055$ and $0.033$ in full-measurement RA correlation, respectively.

\paragraph{Interpolation consistency.}
On-path, the loss improves RA, RD, and RAD correlation by $0.219$, $0.143$, and $0.051$ and precision from $0.162$ to $0.233$, with little change in recall or vehicle coverage.
Off-path, it improves RA PSNR by $1.50$\,dB and SSIM by $0.109$, but lowers RAD correlation ($0.447$ to $0.363$) and F1 ($0.296$ to $0.236$), both still above all baselines.
The loss regularizes intermediate timestamps evaluated on-path, whereas off-path evaluation uses the training timestamps. We retain it to balance temporal interpolation and RA fidelity against off-path RAD fidelity.

\paragraph{PSF modeling.}
We compare the fixed sensor-derived PSF with Gaussian primitives with learned extent and point reflectors with learned PSF bandwidths. 
On-path, the fixed PSF more than doubles detection recall relative to either alternative, improving F1 at similar precision. Off-path, F1 reaches $0.236$, versus $0.074$ for learned extent and $0.124$ for learned bandwidths.
Across both protocols, the fixed PSF also improves object-region correlation in RA, RD, and RAD, despite lower full-RAD PSNR and SSIM.
These results support fixing the sensor response rather than allowing learnable spread to compensate for errors in reflector placement
and reflectivity.
\section{Conclusion, Limitations, and Future Work}
We presented \methodname{}, a method for synthesizing complete RAD measurements of dynamic driving scenes from novel viewpoints. It combines static and motion-tracked point reflectors with a fixed sensor response, separating scene structure from measurement spreading. This formulation improves object reconstruction and detection in both on-path and off-path evaluations and enables sensor-configuration changes without refitting the scene. To our knowledge, it is the first radar NVS method to reconstruct dynamic driving scenes with Doppler measurements.

Several limitations remain. 
Reflector-to-object assignments rely on object annotations and remain fixed during optimization, even as tracks and reflector positions are refined. Jointly refining these assignments could help correct initialization errors, while modeling multipath and speckle noise could improve measurement realism. 
While our formulation supports 3D reconstruction, in this work
we use a ground-plane implementation because the evaluated measurements lack elevation. Future work may evaluate elevation-resolving radar and explore transfer across physical sensors, beyond the processing configurations demonstrated here.

\section*{AI Use Statement}
We used generative AI tools to assist with writing and editing code, drafting and revising manuscript text, and reviewing related literature.
The authors reviewed and revised the AI-assisted manuscript text and code.
We take responsibility for the final content of this work, including all AI-assisted code, text, claims, and artifacts.

\section*{Reproducibility Statement}
The appendix details sensor-specific PSF construction (Sec.~\ref{sec:supp-psf}), scene initialization (Sec.~\ref{sec:supp-scene-initialization}), densification (Sec.~\ref{sec:supp-densification}), and hyperparameter selection, including search ranges and selected values (Sec.~\ref{sec:supp-hyperparameter-selection}).
It also provides evaluation protocols, metric definitions, and additional experimental results (Sec.~\ref{sec:supp-experiments}), together with ablation settings (Sec.~\ref{sec:suppl-ablations}) and configuration-transfer procedures (Sec.~\ref{sec:supp-config-transfer}).
The code, with usage examples and experiment configurations, is available at \codeurl. It includes a deterministic generator for the synthetic dataset described in Sec.~\ref{sec:supp-synthetic}, which serves as a benchmark for off-path radar novel-view synthesis. Videos are available on the \projectpage.

\section*{Acknowledgments} Or Litany acknowledges support from the Israel Science Foundation (grant 624/25) and the Azrieli Foundation Early Career Faculty Fellowship. The authors gratefully acknowledge this support. This research was supported by the Council for Higher Education in Israel under the Moonshot Project.

\bibliography{main}
\bibliographystyle{preprint}

\clearpage
\appendix

\addtocontents{toc}{\protect\contentsline {section}{APPENDIXSTART}{}{}}
\newif\ifinappendix
\inappendixfalse

\begingroup
\makeatletter

\let\clearpage\relax 
\renewcommand{\contentsname}{} 

\setcounter{tocdepth}{2}

\let\oldcontentsline\contentsline
\renewcommand{\contentsline}[4]{%
  \def\temp{#2}%
  \def\marker{APPENDIXSTART}%
  \ifx\temp\marker
    \inappendixtrue
  \else
    \ifinappendix
      \oldcontentsline{#1}{#2}{#3}{#4}%
    \fi
  \fi
}%

\vspace{1.5em}
\noindent{\large\bfseries Contents} 
\vspace{-2.5em}

\renewcommand*\l@section[2]{%
  \ifnum \c@tocdepth >\z@
    \addpenalty\@secpenalty
    \addvspace{-0.35em \@plus\p@}%
    \setlength\@tempdima{1.5em}%
    \begingroup
      \parindent \z@
      \rightskip \@pnumwidth
      \parfillskip -\@pnumwidth
      \leavevmode \bfseries
      \advance\leftskip\@tempdima
      \hskip -\leftskip
      #1\nobreak\hfil
      \nobreak\hb@xt@\@pnumwidth{\hss #2}\par
    \endgroup
  \fi
}

\tableofcontents
\makeatother
\endgroup

\section{Radar Signal Formation}
\label{sec:suppl-signal}

\subsection{Sensor-Specific PSF Construction}
\label{sec:supp-psf}

A localized reflector produces a response across multiple measurement bins because of the radar's acquisition and signal-processing chain. This spreading is a sensor property: changing the reconstructed scene should change the reflectors, not the response applied to them. We therefore construct the sensor response before scene optimization and hold it fixed. For RADIal, we use its known processing operations and calibration table. For Boreas, where these are unavailable, we use an approximate response with fixed parameters selected from sensor specifications and validation data. For the synthetic data we use the same formulation as in RADIal. 

\paragraph{RADIal.}
We construct the kernels from RADIal's reference processing~\citep{radial}. Range and Doppler use Hamming-windowed FFTs. We approximate their normalized power responses with the corresponding analytic Hamming-FFT kernel, evaluated over a five-bin neighborhood.

RADIal separates its transmitters through phase codes that shift their returns by multiples of 16 Doppler FFT bins. From the 256-bin spectrum, we gather the transmitter channels using RADIal's prescribed shifts and retain 16 reduced Doppler offsets. We do not disambiguate individual returns onto the original 256-bin grid. Gathering compensates for the transmitter shifts, leaving the Hamming-FFT response modeled by our Doppler kernel; Doppler indices wrap around the reduced grid.

For azimuth, we compute each source direction's response through RADIal's calibrated beamformer, including its antenna window, then square the magnitude and normalize the peak. This preserves the response's dependence on viewing direction. We use the same elevation slice as RADIal's reference RA processing (index 5, corresponding to $+1^\circ$) for both measurement preprocessing and kernel construction. All kernels remain fixed during scene optimization.

\paragraph{Boreas.}
The processing chain is unavailable, so we approximate the range and azimuth responses using fixed, tapered sinc-squared kernels:
\begin{equation}
h_u(x)=\operatorname{sinc}^2(x/w_u)\,\frac{1+\cos(\pi x/k_u)}{2}\,\mathbf{1}_{|x|\leq k_u},
   \qquad u\in\{R,A\}.
\end{equation}
We use $(w_R,k_R)=(5.7,16)$ and $(w_A,k_A)=(2.26,4)$, in bins. The azimuth width is based on the nominal $1.8^\circ$ beamwidth and $0.9^\circ$ sampling interval; the taper yields an effective width of approximately $1.64^\circ$. The range width is selected by validation reconstruction performance because the available specifications do not provide the range-response width. This is an empirical approximation, held fixed throughout scene optimization. Boreas provides no Doppler axis, so $h_D=1$.

\section{Scene Initialization and Optimization}
\subsection{Scene Initialization Details}
\label{sec:supp-scene-initialization}
\begin{figure}[t]
    \centering
    \includegraphics[width=\linewidth]{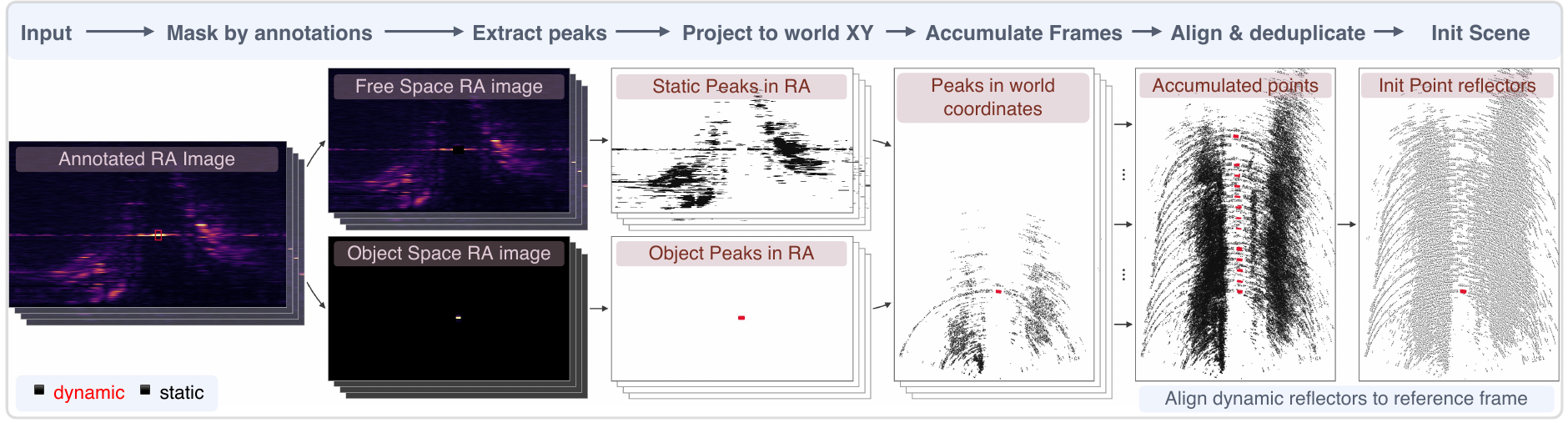}
    \caption{
    \textbf{Point-reflector initialization.}
    Object bounding boxes separate background and object regions.
    Selected returns are projected into world coordinates and
    accumulated across training frames. Dynamic returns are aligned
    by track translation and expressed in object coordinates.
    Deduplication yields the initial static and dynamic reflectors.}
    \label{fig:point-init}
\end{figure}

Figure~\ref{fig:point-init} illustrates the initialization procedure. We initialize the scene using only training-frame measurements and
bounding boxes.

\paragraph{Object association and peak extraction.}
We associate bounding boxes across annotated frames by extrapolating each bounding box from the previous annotated frame using a constant-velocity model and matching it to the current annotations. Each resulting track receives a unique object index.

The annotated bounding box size defines an approximate object footprint, which we dilate by the sensor PSF to obtain its measurement region. The static-background region is the complement of these regions. On RADIal, for the background, we threshold the Doppler-averaged RA map. For each dynamic object, we select the
Doppler bin with the strongest response at its annotated location and threshold that slice within its region. 
For Boreas, we select returns directly from the RA maps, using annotated object regions for dynamic initialization and their complement for the static background.
We project the selected cells onto the ground plane in world coordinates and accumulate
them across frames. Cells in overlapping regions contribute to each corresponding object.

\paragraph{Dynamic-peak alignment.}
Let $\mathbf{z}_i(t)$ be a selected return associated with object
$j$, and let $\mathbf{p}_{0,j}$ be its first training-frame
annotated position. Using the initial track position
$\mathbf{p}_j(t)$, we align the return and express it relative to the object-frame origin:
\begin{equation}
    \boldsymbol{\mu}_i
    = \mathbf{z}_i(t) - \mathbf{p}_j(t).
    \label{eq:dynamic-peak-object-frame}
\end{equation}
This initialization compensates for translation only, approximating
the alignment of returns from turning objects.

\paragraph{Deduplication and parameter initialization.}
We deduplicate static returns on a $0.5$\,m voxel grid and aligned
dynamic returns on a $0.15$\,m grid, separately for each object.
Static reflectors retain their world-space positions and receive
$q_i=0$; dynamic reflectors use $\boldsymbol{\mu}_i$ and receive
$q_i=j$. We initialize the base reflected power to $\alpha_i=0.2$
and all SH coefficients to zero, keeping the DC coefficient fixed.

\subsection{Hyperparameter Selection}
\label{sec:supp-hyperparameter-selection}
We select the interpolation-consistency weight $\lambda_{\mathrm{int}}$ and learning rates for reflector positions, opacities, spherical-harmonic coefficients, and object tracks separately for each dataset using held-out validation data.
We use degree-three SH ($K=16$). All remaining settings are fixed or data-derived. 

We tune five optimization hyperparameters. Each grid contains ten \(1\)-\(2\)-\(5\) logarithmic values. We sweep each hyperparameter independently while holding the remaining four at their pre-study values; the selections are therefore conditional rather than jointly optimal. 

\begin{table}[t]
    \centering
    \caption{Search ranges and selected hyperparameters.}
    \label{tab:hyperparameter-search}
    \begin{tabular}{lccc}
        \toprule
        Hyperparameter & Range & RADIal & Boreas \\
        \midrule
        Interpolation weight \(\lambda_{\mathrm{int}}\)
            & \(10^{-3}\)--\(1\) & \(0.1\) & \(0.01\) \\
        Position LR (\texttt{means\_lr})
            & \(10^{-5}\)--\(10^{-2}\) & \(0.0002\) & \(0.00005\) \\
        Opacity LR (\texttt{opacities\_lr})
            & \(10^{-3}\)--\(1\) & \(0.02\) & \(0.02\) \\
        SH LR (\texttt{sh\_coeffs\_lr})
            & \(10^{-4}\)--\(10^{-1}\) & \(0.01\) & \(0.02\) \\
        Motion LR (\texttt{deform\_lr})
            & \(10^{-4}\)--\(10^{-1}\) & \(0.01\) & \(0.0005\) \\
        \bottomrule
    \end{tabular}
\end{table}

RADIal uses four held-out recordings. Boreas uses four held-out segments of a single annotated recording, each separated from evaluation segments by at least \(1{,}000\) frames. Both datasets use every fifth frame for validation.

We select using a composite of RA correlation, RAD correlation, and peak F1, standardized across candidates within each window and averaged over the four windows. Boreas uses only RA correlation and peak F1, as it has no Doppler dimension. For RADIal, candidates whose object recall falls more than \(0.05\) below the best arm are disqualified.

We also compare per-window rank agreement. It confirms the interpolation-weight selection on RADIal and all five selections on Boreas. For the four RADIal learning rates, rank agreement disagreed or tied; after reviewing these results, we adopted the composite-score winners. Chosen values can be seen in table \ref{tab:hyperparameter-search}.

\subsection{Densification}
\label{sec:supp-densification}
The final model uses residual-guided densification of the static background. Starting at step 1{,}000, we perform densification every 200 steps until the end of training.  At each event, we extract positive-residual peaks within the ego-motion-compensated static-Doppler band, project them into world coordinates, and remove duplicates and candidates within $1\mathrm{m}$ of an existing reflector. We then add the 1{,}000 strongest remaining candidates as static reflectors. We use a residual threshold of 0.05 and limit the complete representation to 800{,}000 reflectors.

We evaluated densification in a 36-configuration sensitivity sweep on the same validation recordings used for hyperparameter selection, disjoint from the evaluation sequences. Performance was stable across the sweep, and we retain its central configuration for every setting.

\section{Additional Experimental Results}
\label{sec:supp-experiments}
\subsection{Evaluation Protocol}
\paragraph{Implementation Details.} 
We optimize \methodname\ and \methodname-static for 10k steps.
For the baselines, we use their published training schedules
on Boreas; on RADIal, we train RadarSplat for 10k steps and
retain the published RadarFields schedule.
All experiments run on a single NVIDIA GeForce RTX 4090.

\paragraph{Reconstruction evaluation.}
\label{sec:suppl-control}
\label{sec:supp-reconstruction-metrics}
We evaluate reconstruction using PSNR, SSIM, LPIPS, and Pearson
correlation $\rho$.
Both datasets are evaluated in their native domain: 
RADIal is evaluated in the linear amplitude domain and Boreas in the sensor's log-compressed domain.
PSNR, SSIM, and correlation are reported over full measurements
and object regions; LPIPS is computed over full RA and RAD
measurements only.

Approximately $97\%$ of RADIal tensor bins lie at the noise floor,
so agreement with the background can dominate full-measurement
scores.
To examine this effect, we evaluate two structureless diagnostic
controls using the same normalization, held-out frames, and
scoring procedure as the reconstruction methods.
\textsc{Const} predicts the median value of each reference RA
map uniformly, while \textsc{Zero} predicts zero everywhere.

\begin{table}[t]
    \centering
    \footnotesize
    \setlength{\tabcolsep}{3pt}
    \renewcommand{\arraystretch}{1.05}
    \caption{\textbf{Structureless controls.}
    Full-measurement RA reconstruction on held-out RADIal frames.
    Mean $\pm$ sample standard deviation across ten sequences;
    best means, including controls, are bold.
    Constant predictions are assigned $\rho=0$ by the scorer.}
    \label{tab:suppl-control} {%
    \begin{tabular}{@{}lcccc@{}}
        \toprule
        Method
        & PSNR [dB] $\uparrow$
        & SSIM $\uparrow$
        & LPIPS $\downarrow$
        & $\rho\uparrow$ \\
        \midrule
        \methodname
        & \meanstd{\mathbf{25.78}}{1.09}
        & \meanstd{0.690}{0.025}
        & \meanstd{\mathbf{0.234}}{0.017}
        & \meanstd{\mathbf{0.612}}{0.031} \\
        \methodname-static
        & \meanstd{25.16}{1.15}
        & \meanstd{0.686}{0.025}
        & \meanstd{0.249}{0.021}
        & \meanstd{0.533}{0.055} \\
        RadarSplat
        & \meanstd{20.94}{0.66}
        & \meanstd{0.050}{0.011}
        & \meanstd{0.533}{0.022}
        & \meanstd{0.291}{0.100} \\
        RadarFields
        & \meanstd{20.38}{0.71}
        & \meanstd{0.059}{0.009}
        & \meanstd{0.569}{0.021}
        & \meanstd{0.055}{0.024} \\
        \midrule
        \textsc{Const}
        & \meanstd{23.71}{1.04}
        & \meanstd{\mathbf{0.723}}{0.023}
        & \meanstd{0.583}{0.025}
        & \meanstd{0.000}{0.000} \\
        \textsc{Zero}
        & \meanstd{20.44}{0.65}
        & \meanstd{0.032}{0.004}
        & \meanstd{0.597}{0.024}
        & \meanstd{0.000}{0.000} \\
        \bottomrule
    \end{tabular}%
    }
\end{table}

Despite containing no scene structure, \textsc{Const} achieves
higher SSIM than every reconstruction method and higher PSNR
than both RadarSplat and RadarFields
(Table~\ref{tab:suppl-control}).
Thus, these metrics alone do not establish that a reconstruction
preserves radar structure.

LPIPS ranks all reconstruction methods above the controls on RA,
although its separation is much larger for \methodname\ than
for the prior baselines.
Correlation measures agreement in spatial variation independently
of an affine intensity transformation; it therefore complements
PSNR without measuring intensity calibration.
For constant predictions, where Pearson correlation is undefined,
the scorer assigns zero.

We consequently interpret PSNR and SSIM alongside correlation
and LPIPS, and report object-region scores to expose errors
that full-measurement averages can conceal.
Detection evaluation further tests whether synthesized
measurements preserve usable radar returns.

\paragraph{Detection evaluation.}
\label{sec:supp-detection-metrics}
We apply the CFAR detector from RADIal’s original implementation, with unchanged settings, to both measured and synthesized tensors. This evaluates compatibility with the dataset’s native detection pipeline without detector-specific tuning.
We then evaluate the resulting point clouds using metrics from RadarGen~\citep{radargen}, with return power in place of calibrated RCS.
Entire-area metrics comprise CD for spatial localization,
CD-Full for joint distance in normalized position, Doppler,
and power, and IoU@1\,m for spatial overlap.
DA precision, recall, and F1 require agreement in position
and both radar attributes.
Maximum Mean Discrepancy (MMD) measures distributional agreement
separately for location, Doppler, and power, using five
multi-scale RBF kernels and fixed-seed subsampling to at most
2,000 points.
Doppler comparisons account for periodicity, with the DA
Doppler threshold set to $10\%$ of RADIal's Doppler period.

Foreground metrics are averaged over annotated boxes.
Hit rate measures the fraction of boxes containing reference
detections that also contain predicted detections;
density similarity is the ratio of the smaller detection
count to the larger, with value one when both are zero.
Both penalize missing returns in reference-occupied boxes.
We additionally report the number of boxes containing
predictions but no reference detections (FP boxes).

We omit conditional foreground CD, CD-Full, and MMD:
the baselines miss most objects, leaving these scores
undefined or evaluated on a small detected subset.
They therefore cannot support a comparison across methods.
We instead report foreground hit rate and density similarity,
which penalize missed objects.
Tables report means and sample standard deviations across
sequences, or scene--offset pairs for synthetic evaluation.

\subsection{On-Path Evaluation}
\label{sec:supp-onpath}
\begin{figure*}[p]
    \centering
    \includegraphics[
        width=\textwidth,
        height=0.86\textheight,
        keepaspectratio
    ]{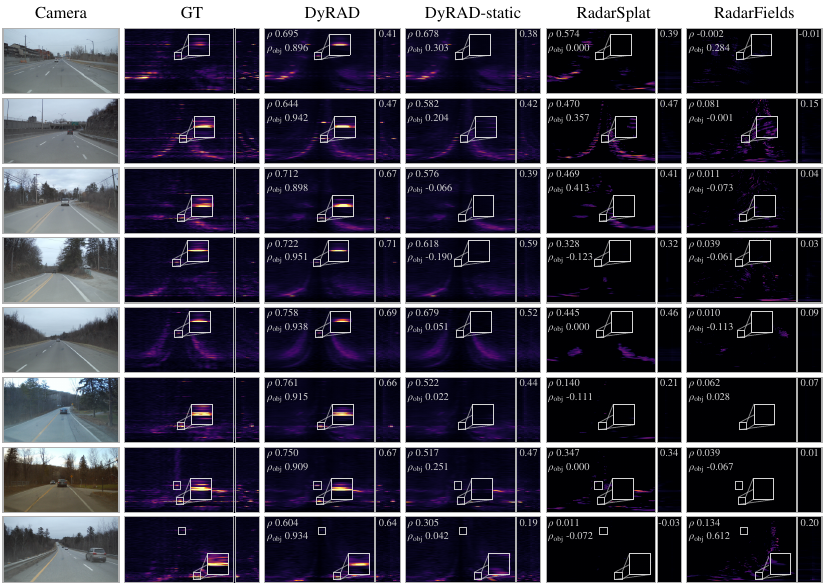}
    \caption{\textbf{Additional on-path examples on RADIal.}
    Each row compares a held-out measurement with the corresponding reconstructions, showing RA (left) and RD (right) projections within each panel. Displayed correlations are computed separately for each projection. White boxes mark objects; insets enlarge selected RA object regions. \methodname\ preserves object returns that the static variant and baselines miss, even where the surrounding scene is reconstructed similarly.}
    \label{fig:supp_onpath_radial}
\end{figure*}

\begin{figure*}[t]
    \centering
    \includegraphics[width=\textwidth]{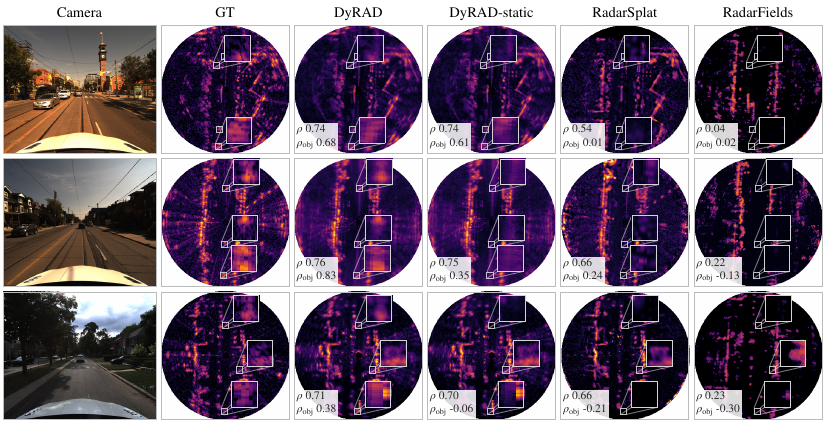}
    \caption{\textbf{On-path examples on Boreas.}
    Held-out RA measurements and reconstructions displayed
    in Cartesian bird's-eye view.
    White boxes mark annotated objects; insets enlarge selected object regions. \methodname\ preserves object returns that the static variant and baselines miss, even where the surrounding scene is reconstructed similarly.}
    \label{fig:supp_onpath_boreas}
\end{figure*}

Tables~\ref{tab:supp-onpath-reconstruction} and~\ref{tab:supp-onpath-detection} provide the full reconstruction and detection results for the on-path evaluation described in Sec.~\ref{sec:nvs}.
Figures~\ref{fig:supp_onpath_radial} and~\ref{fig:supp_onpath_boreas} show qualitative comparisons on RADIal and Boreas, respectively.

\begin{table*}[t]
    \centering
    \footnotesize
    \setlength{\tabcolsep}{3pt}
    \renewcommand{\arraystretch}{1.05}
    \caption{\textbf{On-path reconstruction: full results.}
    Mean $\pm$ sample standard deviation across ten RADIal
    sequences and three Boreas sequences.
    Best means within each projection are bold.
    $^{*}$ denotes the ego-Doppler lift.
    LPIPS is not reported for RD.}
    \label{tab:supp-onpath-reconstruction}
    \resizebox{\textwidth}{!}{%
    \begin{tabular}{lccccccc}
        \toprule
        & \multicolumn{4}{c}{Full}
        & \multicolumn{3}{c}{Object} \\
        \cmidrule(lr){2-5}\cmidrule(lr){6-8}
        Method
        & $\rho\uparrow$
        & PSNR [dB] $\uparrow$
        & SSIM $\uparrow$
        & LPIPS $\downarrow$
        & $\rho\uparrow$
        & PSNR [dB] $\uparrow$
        & SSIM $\uparrow$ \\
        \midrule
        \multicolumn{8}{l}{\emph{RADIal --- RAD reconstruction}} \\
        RadarSplat$^{*}$
        & \meanstd{0.068}{0.028}
        & \meanstd{23.56}{0.85}
        & \meanstd{0.215}{0.006}
        & \meanstd{0.634}{0.008}
        & \meanstd{-0.010}{0.057}
        & \meanstd{13.54}{1.22}
        & \meanstd{0.098}{0.021} \\
        RadarFields$^{*}$
        & \meanstd{0.015}{0.009}
        & \meanstd{23.13}{1.23}
        & \meanstd{0.203}{0.012}
        & \meanstd{0.637}{0.007}
        & \meanstd{-0.011}{0.051}
        & \meanstd{13.40}{1.28}
        & \meanstd{0.088}{0.021} \\
        \methodname-static
        & \meanstd{0.213}{0.054}
        & \meanstd{25.67}{0.95}
        & \meanstd{0.498}{0.021}
        & \meanstd{0.279}{0.026}
        & \meanstd{-0.013}{0.072}
        & \meanstd{13.76}{1.24}
        & \meanstd{0.199}{0.035} \\
        \methodname
        & \meanstd{\mathbf{0.272}}{0.062}
        & \meanstd{\mathbf{25.88}}{0.95}
        & \meanstd{\mathbf{0.502}}{0.022}
        & \meanstd{\mathbf{0.275}}{0.025}
        & \meanstd{\mathbf{0.658}}{0.070}
        & \meanstd{\mathbf{16.97}}{1.06}
        & \meanstd{\mathbf{0.374}}{0.028} \\
        \midrule
        \multicolumn{8}{l}{\emph{RADIal --- RA reconstruction}} \\
        RadarSplat
        & \meanstd{0.291}{0.100}
        & \meanstd{20.94}{0.66}
        & \meanstd{0.050}{0.011}
        & \meanstd{0.533}{0.022}
        & \meanstd{-0.005}{0.024}
        & \meanstd{8.08}{1.45}
        & \meanstd{0.008}{0.004} \\
        RadarFields
        & \meanstd{0.055}{0.024}
        & \meanstd{20.38}{0.71}
        & \meanstd{0.059}{0.009}
        & \meanstd{0.569}{0.021}
        & \meanstd{-0.004}{0.063}
        & \meanstd{8.18}{1.43}
        & \meanstd{0.019}{0.007} \\
        \methodname-static
        & \meanstd{0.533}{0.055}
        & \meanstd{25.16}{1.15}
        & \meanstd{0.686}{0.025}
        & \meanstd{0.249}{0.021}
        & \meanstd{0.021}{0.088}
        & \meanstd{8.70}{1.56}
        & \meanstd{0.232}{0.058} \\
        \methodname
        & \meanstd{\mathbf{0.612}}{0.031}
        & \meanstd{\mathbf{25.78}}{1.09}
        & \meanstd{\mathbf{0.690}}{0.025}
        & \meanstd{\mathbf{0.234}}{0.017}
        & \meanstd{\mathbf{0.827}}{0.087}
        & \meanstd{\mathbf{14.93}}{1.89}
        & \meanstd{\mathbf{0.568}}{0.097} \\
        \midrule
        \multicolumn{8}{l}{\emph{RADIal --- RD reconstruction}} \\
        RadarSplat$^{*}$
        & \meanstd{0.303}{0.090}
        & \meanstd{21.57}{0.93}
        & \meanstd{0.102}{0.026}
        & --
        & \meanstd{0.073}{0.104}
        & \meanstd{16.81}{2.02}
        & \meanstd{0.063}{0.015} \\
        RadarFields$^{*}$
        & \meanstd{0.131}{0.062}
        & \meanstd{21.44}{0.93}
        & \meanstd{0.129}{0.036}
        & --
        & \meanstd{0.019}{0.063}
        & \meanstd{16.87}{2.06}
        & \meanstd{0.098}{0.043} \\
        \methodname-static
        & \meanstd{0.375}{0.064}
        & \meanstd{24.78}{1.68}
        & \meanstd{0.611}{0.038}
        & --
        & \meanstd{0.073}{0.084}
        & \meanstd{18.56}{2.39}
        & \meanstd{0.438}{0.111} \\
        \methodname
        & \meanstd{\mathbf{0.476}}{0.067}
        & \meanstd{\mathbf{25.33}}{1.70}
        & \meanstd{\mathbf{0.620}}{0.040}
        & --
        & \meanstd{\mathbf{0.564}}{0.100}
        & \meanstd{\mathbf{20.79}}{1.95}
        & \meanstd{\mathbf{0.558}}{0.086} \\
        \midrule
        \multicolumn{8}{l}{\emph{Boreas --- RA reconstruction}} \\
        RadarSplat
        & \meanstd{0.666}{0.036}
        & \meanstd{23.92}{1.07}
        & \meanstd{0.446}{0.044}
        & \meanstd{0.411}{0.005}
        & \meanstd{0.178}{0.249}
        & \meanstd{19.61}{3.23}
        & \meanstd{0.273}{0.121} \\
        RadarFields
        & \meanstd{0.279}{0.078}
        & \meanstd{19.96}{0.84}
        & \meanstd{0.147}{0.011}
        & \meanstd{0.518}{0.016}
        & \meanstd{0.016}{0.128}
        & \meanstd{16.99}{1.92}
        & \meanstd{0.041}{0.041} \\
        \methodname-static
        & \meanstd{0.749}{0.018}
        & \meanstd{25.31}{0.90}
        & \meanstd{0.482}{0.037}
        & \meanstd{\mathbf{0.304}}{0.011}
        & \meanstd{0.278}{0.217}
        & \meanstd{21.13}{2.89}
        & \meanstd{0.372}{0.087} \\
        \methodname
        & \meanstd{\mathbf{0.752}}{0.017}
        & \meanstd{\mathbf{25.36}}{0.90}
        & \meanstd{\mathbf{0.484}}{0.038}
        & \meanstd{0.305}{0.010}
        & \meanstd{\mathbf{0.622}}{0.130}
        & \meanstd{\mathbf{24.78}}{0.54}
        & \meanstd{\mathbf{0.565}}{0.129} \\
        \bottomrule
    \end{tabular}%
    }
\end{table*}

\begin{table*}[t]
    \centering
    \footnotesize
    \setlength{\tabcolsep}{4pt}
    \renewcommand{\arraystretch}{1.05}
    \caption{\textbf{On-path detection on RADIal: full results.}
    Mean $\pm$ sample standard deviation across ten sequences.
    Best means, including ties, are bold.
    $^{*}$ denotes the ego-Doppler lift.}
    \label{tab:supp-onpath-detection}
    \resizebox{\textwidth}{!}{%
    \begin{tabular}{lcccccc}
        \toprule
        & \multicolumn{6}{c}{Entire area} \\
        \cmidrule(lr){2-7}
        Method
        & CD [m] $\downarrow$
        & CD-Full $\downarrow$
        & IoU@1\,m $\uparrow$
        & Precision $\uparrow$
        & Recall $\uparrow$
        & F1 $\uparrow$ \\
        \midrule
        RadarSplat$^{*}$
        & \meanstd{6.53}{5.82}
        & \meanstd{0.1668}{0.0758}
        & \meanstd{0.141}{0.066}
        & \meanstd{0.084}{0.036}
        & \meanstd{0.087}{0.056}
        & \meanstd{0.073}{0.048} \\
        RadarFields$^{*}$
        & \meanstd{6.97}{0.96}
        & \meanstd{0.1612}{0.0521}
        & \meanstd{0.037}{0.012}
        & \meanstd{0.015}{0.006}
        & \meanstd{0.019}{0.010}
        & \meanstd{0.016}{0.007} \\
        \methodname-static
        & \meanstd{4.71}{1.29}
        & \meanstd{0.0783}{0.0117}
        & \meanstd{0.188}{0.043}
        & \meanstd{0.169}{0.055}
        & \meanstd{0.099}{0.034}
        & \meanstd{0.123}{0.041} \\
        \methodname
        & \meanstd{\mathbf{4.01}}{1.02}
        & \meanstd{\mathbf{0.0689}}{0.0092}
        & \meanstd{\mathbf{0.238}}{0.040}
        & \meanstd{\mathbf{0.233}}{0.051}
        & \meanstd{\mathbf{0.149}}{0.037}
        & \meanstd{\mathbf{0.179}}{0.041} \\
        \midrule
        & \multicolumn{3}{c}{Entire area}
        & \multicolumn{3}{c}{Foreground} \\
        \cmidrule(lr){2-4}\cmidrule(lr){5-7}
        Method
        & MMD location $\downarrow$
        & MMD Doppler $\downarrow$
        & MMD power $\downarrow$
        & Hit rate $\uparrow$
        & Density sim. $\uparrow$
        & FP boxes $\downarrow$ \\
        \midrule
        RadarSplat$^{*}$
        & \meanstd{0.9263}{0.4340}
        & \meanstd{1.1820}{0.8409}
        & \meanstd{5.6300}{0.6544}
        & \meanstd{0.036}{0.067}
        & \meanstd{0.048}{0.048}
        & \meanstd{\mathbf{0.0}}{0.0} \\
        RadarFields$^{*}$
        & \meanstd{0.5385}{0.1566}
        & \meanstd{1.3328}{0.6162}
        & \meanstd{4.7832}{0.8515}
        & \meanstd{0.269}{0.245}
        & \meanstd{0.153}{0.121}
        & \meanstd{\mathbf{0.0}}{0.0} \\
        \methodname-static
        & \meanstd{0.5842}{0.2191}
        & \meanstd{0.2386}{0.1191}
        & \meanstd{0.7471}{0.2970}
        & \meanstd{0.070}{0.083}
        & \meanstd{0.047}{0.057}
        & \meanstd{\mathbf{0.0}}{0.0} \\
        \methodname
        & \meanstd{\mathbf{0.4208}}{0.1425}
        & \meanstd{\mathbf{0.1657}}{0.0741}
        & \meanstd{\mathbf{0.5933}}{0.2082}
        & \meanstd{\mathbf{0.907}}{0.086}
        & \meanstd{\mathbf{0.583}}{0.099}
        & \meanstd{0.3}{0.5} \\
        \bottomrule
    \end{tabular}%
    }
\end{table*}

\subsection{Real-Data Off-Path Evaluation}
\label{sec:supp-offpath}

\begin{figure*}[t]
    \centering
    \includegraphics[width=\textwidth]
        {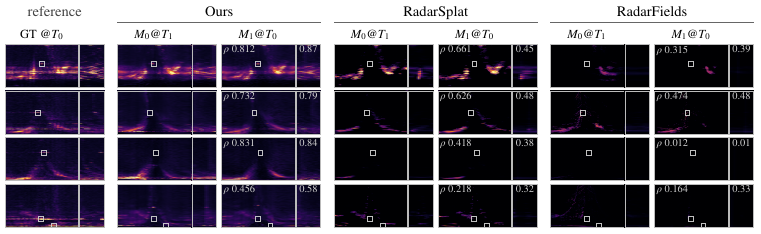}
    \caption{\textbf{Off-path rendering and round-trip reconstruction on RADIal.} Each panel shows RA (left) and RD (right) projections. For each method, $M_0@T_1$ shows measurements synthesized along the trajectory shifted laterally by $2\,\mathrm{m}$. A new model is fitted to these renders and evaluated at the original poses ($M_1@T_0$), against the real measurements (GT@$T_0$). White boxes mark annotated vehicles. The values in the return panels report RA and RD correlations with the reference, respectively.}
    \label{fig:supp_cycle_radial}
\end{figure*}

\begin{figure*}[t]
    \centering
    \includegraphics[width=\textwidth]
        {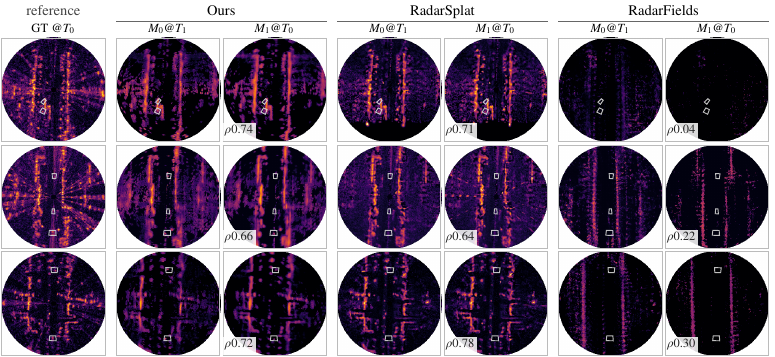}
    \caption{\textbf{Off-path rendering and round-trip reconstruction on Boreas.}
    RA measurements shown in Cartesian bird's-eye view.
    For each method, $M_0@T_1$ shows the initial model rendered
    along the trajectory shifted laterally by $2\,\mathrm{m}$.
    A new model is fitted to these renders and evaluated at
    the original poses ($M_1@T_0$), against the real
    measurements (GT@$T_0$).
    White boxes mark annotated vehicles; $\rho$ denotes
    correlation of the round-trip reconstruction with the reference.}
    \label{fig:supp_cycle_boreas}
\end{figure*}

Tables~\ref{tab:supp_offpath_reconstruction} and~\ref{tab:supp_offpath_detection} provide the full reconstruction and detection results for the real-data off-path protocol described in Sec.~\ref{sec:nvs}.
Figures~\ref{fig:supp_cycle_radial} and~\ref{fig:supp_cycle_boreas} illustrate the displaced renders and subsequent reconstructions at the original poses.
On RADIal, RadarFields produces a near-constant reconstruction on one sequence and an all-zero reconstruction on another; both are included in the reported averages.

\begin{table*}[t]
    \centering
    \footnotesize
    \setlength{\tabcolsep}{3pt}
    \renewcommand{\arraystretch}{1.05}
    \caption{\textbf{Real-data off-path reconstruction: full results.}
    Results at the original poses after a $2\,\mathrm{m}$
    lateral-shift round trip.
    Mean $\pm$ sample standard deviation across ten RADIal
    sequences and three Boreas sequences; best means are bold.
    $^{*}$ denotes the ego-Doppler lift.
    LPIPS is not reported for RD.}
    \label{tab:supp_offpath_reconstruction}
    \resizebox{\textwidth}{!}{%
    \begin{tabular}{lccccccc}
        \toprule
        & \multicolumn{4}{c}{Full}
        & \multicolumn{3}{c}{Object} \\
        \cmidrule(lr){2-5}\cmidrule(lr){6-8}
        Method
        & $\rho\uparrow$
        & PSNR [dB] $\uparrow$
        & SSIM $\uparrow$
        & LPIPS $\downarrow$
        & $\rho\uparrow$
        & PSNR [dB] $\uparrow$
        & SSIM $\uparrow$ \\
        \midrule
        \multicolumn{8}{l}{\emph{RADIal --- RAD reconstruction}} \\
        RadarSplat$^{*}$
        & \meanstd{0.076}{0.032}
        & \meanstd{23.92}{0.85}
        & \meanstd{0.217}{0.006}
        & \meanstd{0.634}{0.008}
        & \meanstd{-0.004}{0.053}
        & \meanstd{13.57}{1.32}
        & \meanstd{0.099}{0.022} \\
        RadarFields$^{*}$
        & \meanstd{0.023}{0.018}
        & \meanstd{24.43}{0.75}
        & \meanstd{0.217}{0.006}
        & \meanstd{0.638}{0.008}
        & \meanstd{-0.011}{0.048}
        & \meanstd{13.48}{1.20}
        & \meanstd{0.099}{0.023} \\
        \methodname-static
        & \meanstd{0.335}{0.054}
        & \meanstd{26.41}{0.73}
        & \meanstd{0.492}{0.012}
        & \meanstd{0.271}{0.014}
        & \meanstd{0.056}{0.064}
        & \meanstd{13.86}{1.35}
        & \meanstd{0.194}{0.035} \\
        \methodname
        & \meanstd{\mathbf{0.363}}{0.052}
        & \meanstd{\mathbf{26.53}}{0.74}
        & \meanstd{\mathbf{0.498}}{0.012}
        & \meanstd{\mathbf{0.267}}{0.015}
        & \meanstd{\mathbf{0.536}}{0.077}
        & \meanstd{\mathbf{15.98}}{1.36}
        & \meanstd{\mathbf{0.344}}{0.041} \\
        \midrule
        \multicolumn{8}{l}{\emph{RADIal --- RA reconstruction}} \\
        RadarSplat
        & \meanstd{0.357}{0.125}
        & \meanstd{21.15}{0.73}
        & \meanstd{0.051}{0.012}
        & \meanstd{0.536}{0.023}
        & \meanstd{0.012}{0.017}
        & \meanstd{8.12}{1.63}
        & \meanstd{0.010}{0.006} \\
        RadarFields
        & \meanstd{0.104}{0.067}
        & \meanstd{20.57}{0.65}
        & \meanstd{0.038}{0.006}
        & \meanstd{0.572}{0.027}
        & \meanstd{0.015}{0.029}
        & \meanstd{8.14}{1.61}
        & \meanstd{0.010}{0.005} \\
        \methodname-static
        & \meanstd{0.595}{0.056}
        & \meanstd{25.56}{0.99}
        & \meanstd{0.652}{0.019}
        & \meanstd{0.218}{0.015}
        & \meanstd{0.114}{0.066}
        & \meanstd{8.76}{1.76}
        & \meanstd{0.215}{0.068} \\
        \methodname
        & \meanstd{\mathbf{0.661}}{0.035}
        & \meanstd{\mathbf{26.15}}{0.95}
        & \meanstd{\mathbf{0.665}}{0.023}
        & \meanstd{\mathbf{0.212}}{0.015}
        & \meanstd{\mathbf{0.807}}{0.057}
        & \meanstd{\mathbf{14.14}}{1.66}
        & \meanstd{\mathbf{0.517}}{0.089} \\
        \midrule
        \multicolumn{8}{l}{\emph{RADIal --- RD reconstruction}} \\
        RadarSplat$^{*}$
        & \meanstd{0.301}{0.115}
        & \meanstd{21.50}{0.94}
        & \meanstd{0.091}{0.023}
        & --
        & \meanstd{0.063}{0.098}
        & \meanstd{16.93}{2.05}
        & \meanstd{0.058}{0.014} \\
        RadarFields$^{*}$
        & \meanstd{0.086}{0.066}
        & \meanstd{21.12}{0.96}
        & \meanstd{0.066}{0.022}
        & --
        & \meanstd{0.021}{0.042}
        & \meanstd{16.85}{2.05}
        & \meanstd{0.043}{0.020} \\
        \methodname-static
        & \meanstd{0.633}{0.059}
        & \meanstd{26.35}{1.24}
        & \meanstd{0.653}{0.021}
        & --
        & \meanstd{0.512}{0.152}
        & \meanstd{19.74}{2.01}
        & \meanstd{0.505}{0.057} \\
        \methodname
        & \meanstd{\mathbf{0.653}}{0.048}
        & \meanstd{\mathbf{26.62}}{1.26}
        & \meanstd{\mathbf{0.661}}{0.022}
        & --
        & \meanstd{\mathbf{0.584}}{0.094}
        & \meanstd{\mathbf{20.56}}{1.97}
        & \meanstd{\mathbf{0.551}}{0.053} \\
        \midrule
        \multicolumn{8}{l}{\emph{Boreas --- RA reconstruction}} \\
        RadarSplat
        & \meanstd{0.689}{0.042}
        & \meanstd{24.29}{1.16}
        & \meanstd{\mathbf{0.490}}{0.040}
        & \meanstd{\mathbf{0.415}}{0.004}
        & \meanstd{0.211}{0.237}
        & \meanstd{19.25}{3.51}
        & \meanstd{0.289}{0.123} \\
        RadarFields
        & \meanstd{0.538}{0.081}
        & \meanstd{21.09}{1.28}
        & \meanstd{0.466}{0.044}
        & \meanstd{0.550}{0.007}
        & \meanstd{0.095}{0.123}
        & \meanstd{16.54}{2.30}
        & \meanstd{0.296}{0.069} \\
        \methodname-static
        & \meanstd{0.741}{0.007}
        & \meanstd{24.45}{0.62}
        & \meanstd{0.359}{0.023}
        & \meanstd{0.462}{0.024}
        & \meanstd{0.261}{0.215}
        & \meanstd{20.02}{2.54}
        & \meanstd{0.296}{0.065} \\
        \methodname
        & \meanstd{\mathbf{0.744}}{0.009}
        & \meanstd{\mathbf{24.51}}{0.62}
        & \meanstd{0.360}{0.020}
        & \meanstd{0.461}{0.023}
        & \meanstd{\mathbf{0.641}}{0.165}
        & \meanstd{\mathbf{24.19}}{1.39}
        & \meanstd{\mathbf{0.519}}{0.180} \\
        \bottomrule
    \end{tabular}%
    }
\end{table*}

\begin{table*}[t]
    \centering
    \footnotesize
    \setlength{\tabcolsep}{4pt}
    \renewcommand{\arraystretch}{1.05}
    \caption{\textbf{Real-data off-path detection on RADIal: full results.}
    Results at the original poses after a $2\,\mathrm{m}$
    lateral-shift round trip.
    Mean $\pm$ sample standard deviation across ten sequences;
    best means are bold.
    $^{*}$ denotes the ego-Doppler lift.}
    \label{tab:supp_offpath_detection}

    \resizebox{\textwidth}{!}{%
    \begin{tabular}{lcccccc}
        \toprule
        & \multicolumn{6}{c}{Entire area} \\
        \cmidrule(lr){2-7}
        Method
        & CD [m] $\downarrow$
        & CD-Full $\downarrow$
        & IoU@1\,m $\uparrow$
        & Precision $\uparrow$
        & Recall $\uparrow$
        & F1 $\uparrow$ \\
        \midrule
        RadarSplat$^{*}$
        & \meanstd{6.35}{4.75}
        & \meanstd{0.1699}{0.0702}
        & \meanstd{0.155}{0.076}
        & \meanstd{0.085}{0.042}
        & \meanstd{0.093}{0.061}
        & \meanstd{0.077}{0.050} \\
        RadarFields$^{*}$
        & \meanstd{9.19}{6.07}
        & \meanstd{0.2211}{0.0962}
        & \meanstd{0.072}{0.045}
        & \meanstd{0.198}{0.353}
        & \meanstd{0.033}{0.026}
        & \meanstd{0.027}{0.020} \\
        \methodname-static
        & \meanstd{3.47}{0.61}
        & \meanstd{0.0681}{0.0101}
        & \meanstd{0.239}{0.036}
        & \meanstd{0.221}{0.051}
        & \meanstd{0.205}{0.036}
        & \meanstd{0.210}{0.042} \\
        \methodname
        & \meanstd{\mathbf{3.25}}{0.57}
        & \meanstd{\mathbf{0.0629}}{0.0085}
        & \meanstd{\mathbf{0.268}}{0.034}
        & \meanstd{\mathbf{0.251}}{0.049}
        & \meanstd{\mathbf{0.227}}{0.034}
        & \meanstd{\mathbf{0.236}}{0.040} \\
        \bottomrule
    \end{tabular}%
    }

    \vspace{4pt}

    \resizebox{\textwidth}{!}{%
    \begin{tabular}{lccccccc}
        \toprule
        & \multicolumn{3}{c}{Entire area}
        & \multicolumn{4}{c}{Foreground} \\
        \cmidrule(lr){2-4}\cmidrule(lr){5-8}
        Method
        & MMD location $\downarrow$
        & MMD Doppler $\downarrow$
        & MMD power $\downarrow$
        & Hit rate $\uparrow$
        & Miss rate $\downarrow$
        & Density sim. $\uparrow$
        & FP boxes $\downarrow$ \\
        \midrule
        RadarSplat$^{*}$
        & \meanstd{1.0231}{0.5973}
        & \meanstd{1.2247}{0.7808}
        & \meanstd{5.7598}{0.6720}
        & \meanstd{0.064}{0.065}
        & \meanstd{0.936}{0.065}
        & \meanstd{0.058}{0.038}
        & \meanstd{\mathbf{0.0}}{0.0} \\
        RadarFields$^{*}$
        & \meanstd{1.0864}{0.4843}
        & \meanstd{1.9832}{1.1824}
        & \meanstd{6.1715}{0.7454}
        & \meanstd{0.181}{0.183}
        & \meanstd{0.819}{0.183}
        & \meanstd{0.108}{0.097}
        & \meanstd{0.3}{0.5} \\
        \methodname-static
        & \meanstd{0.3965}{0.1717}
        & \meanstd{0.1867}{0.0857}
        & \meanstd{1.3521}{0.2469}
        & \meanstd{0.209}{0.144}
        & \meanstd{0.791}{0.144}
        & \meanstd{0.085}{0.073}
        & \meanstd{0.2}{0.6} \\
        \methodname
        & \meanstd{\mathbf{0.3534}}{0.1606}
        & \meanstd{\mathbf{0.1749}}{0.0822}
        & \meanstd{\mathbf{1.2026}}{0.2107}
        & \meanstd{\mathbf{0.917}}{0.101}
        & \meanstd{\mathbf{0.083}}{0.101}
        & \meanstd{\mathbf{0.521}}{0.111}
        & \meanstd{1.1}{1.2} \\
        \bottomrule
    \end{tabular}%
    }
\end{table*}

\subsection{Synthetic Off-Path Evaluation}
\label{sec:supp-synthetic}
\paragraph{Scene generation.}
We construct five scenes with crossing traffic, a curved road, sparse roadside structure, a dual carriageway, and an urban canyon.
Static structures and vehicles comprise world-frame reflectors with positions, velocities, reflectivities, and surface normals.
The scene evolves independently of the sensor trajectory, allowing the same world state to be rendered from different poses at each timestamp.
Each sequence contains 40 frames at $0.1\,\mathrm{s}$ intervals.

The generator produces RADIal-format tensors with 512 range bins, 751 azimuth bins, and 16 Doppler bins, matching the real data; both are cropped identically to 447 range bins before training and scoring. These are generated together with sensor poses, ego velocities, and view-specific vehicle annotations.
Measurements use analytic sensor responses in the amplitude domain and log-normal clutter.
Doppler is wrapped to RADIal's $1.7968\,\mathrm{m/s}$ period; real DDMA demultiplexing artifacts are not reproduced.
The benchmark therefore provides ground truth under an idealized sensor model.
We plan to publicly release the synthetic scenes and their off-path ground-truth measurements to support reuse of the benchmark.

\paragraph{Evaluation protocol.}
Each method is fitted to the base trajectory and evaluated without refitting at five displaced trajectories: lateral offsets of $-1.75$, $+1.75$, and $+3.5\,\mathrm{m}$,
and yaw offsets of $+5^\circ$ and $+10^\circ$. 
Results are pooled over the 25 scene--offset pairs; the reported standard deviation describes variation across scenes and views, not training seeds.

\paragraph{Full results.}
Tables~\ref{tab:supp-synthetic} and~\ref{tab:supp-synthetic-detection} provide the full reconstruction and detection results.
For object-region RD metrics, the evaluation mask includes all Doppler bins at ranges occupied by annotated vehicles; collapsing azimuth therefore also includes background returns at those ranges.
The \projectpage{} provides videos of synthetic off-path sequences for all methods.
\begin{table*}[t]
    \centering
    \footnotesize
    \setlength{\tabcolsep}{3pt}
    \renewcommand{\arraystretch}{1.05}
    \caption{\textbf{Synthetic off-path reconstruction.}
    Mean $\pm$ sample standard deviation across 25 scene--offset pairs;
    best displayed means are bold.
    $^{*}$ denotes the wrap-aware static ego-Doppler lift of RA predictions.
    LPIPS is reported only for full RA and RAD.}
    \label{tab:supp-synthetic}
    \ablationfit{%
    \begin{tabular}{@{}lccccccc@{}}
    \toprule
    & \multicolumn{4}{c}{Full}
    & \multicolumn{3}{c}{Object} \\
    \cmidrule(lr){2-5}\cmidrule(lr){6-8}
    Method
    & $\rho\uparrow$
    & PSNR [dB] $\uparrow$
    & SSIM $\uparrow$
    & LPIPS $\downarrow$
    & $\rho\uparrow$
    & PSNR [dB] $\uparrow$
    & SSIM $\uparrow$ \\
    \midrule
    \multicolumn{8}{l}{\emph{RAD reconstruction}} \\
    RadarSplat$^{*}$
    & \meanstd{0.171}{0.085}
    & \meanstd{22.17}{1.15}
    & \meanstd{0.080}{0.005}
    & \meanstd{0.678}{0.013}
    & \meanstd{0.085}{0.179}
    & \meanstd{10.24}{1.20}
    & \meanstd{0.027}{0.022} \\
    RadarFields$^{*}$
    & \meanstd{0.035}{0.024}
    & \meanstd{22.39}{1.18}
    & \meanstd{0.075}{0.001}
    & \meanstd{0.703}{0.013}
    & \meanstd{0.086}{0.171}
    & \meanstd{10.44}{0.99}
    & \meanstd{0.022}{0.005} \\
    \methodname-static
    & \meanstd{0.297}{0.158}
    & \meanstd{23.94}{0.58}
    & \meanstd{0.290}{0.014}
    & \meanstd{0.522}{0.047}
    & \meanstd{0.161}{0.270}
    & \meanstd{11.14}{1.32}
    & \meanstd{0.121}{0.045} \\
    \methodname
    & \meanstd{\mathbf{0.324}}{0.153}
    & \meanstd{\mathbf{24.06}}{0.55}
    & \meanstd{\mathbf{0.291}}{0.014}
    & \meanstd{\mathbf{0.516}}{0.045}
    & \meanstd{\mathbf{0.356}}{0.151}
    & \meanstd{\mathbf{11.89}}{1.49}
    & \meanstd{\mathbf{0.215}}{0.073} \\
    \midrule
    \multicolumn{8}{l}{\emph{RA reconstruction}} \\
    RadarSplat
    & \meanstd{0.467}{0.195}
    & \meanstd{26.67}{5.19}
    & \meanstd{0.333}{0.224}
    & \meanstd{0.300}{0.269}
    & \meanstd{0.070}{0.060}
    & \meanstd{13.69}{5.90}
    & \meanstd{0.092}{0.078} \\
    RadarFields
    & \meanstd{0.051}{0.053}
    & \meanstd{25.30}{4.09}
    & \meanstd{0.313}{0.213}
    & \meanstd{0.360}{0.270}
    & \meanstd{-0.005}{0.018}
    & \meanstd{13.22}{5.49}
    & \meanstd{0.028}{0.026} \\
    \methodname-static
    & \meanstd{0.514}{0.239}
    & \meanstd{28.38}{4.45}
    & \meanstd{0.635}{0.215}
    & \meanstd{0.255}{0.216}
    & \meanstd{0.106}{0.143}
    & \meanstd{13.80}{5.72}
    & \meanstd{0.103}{0.071} \\
    \methodname
    & \meanstd{\mathbf{0.574}}{0.227}
    & \meanstd{\mathbf{28.85}}{4.54}
    & \meanstd{\mathbf{0.640}}{0.215}
    & \meanstd{\mathbf{0.247}}{0.217}
    & \meanstd{\mathbf{0.593}}{0.237}
    & \meanstd{\mathbf{17.39}}{3.38}
    & \meanstd{\mathbf{0.542}}{0.126} \\
    \midrule
    \multicolumn{8}{l}{\emph{RD reconstruction}} \\
    RadarSplat$^{*}$
    & \meanstd{0.299}{0.154}
    & \meanstd{21.78}{4.38}
    & \meanstd{0.142}{0.094}
    & --
    & \meanstd{0.209}{0.140}
    & \meanstd{20.42}{5.11}
    & \meanstd{0.091}{0.065} \\
    RadarFields$^{*}$
    & \meanstd{0.057}{0.046}
    & \meanstd{21.38}{4.20}
    & \meanstd{0.124}{0.088}
    & --
    & \meanstd{0.034}{0.034}
    & \meanstd{20.19}{4.94}
    & \meanstd{0.066}{0.046} \\
    \methodname-static
    & \meanstd{0.492}{0.185}
    & \meanstd{23.90}{4.04}
    & \meanstd{0.530}{0.189}
    & --
    & \meanstd{0.333}{0.127}
    & \meanstd{22.00}{4.93}
    & \meanstd{0.389}{0.132} \\
    \methodname
    & \meanstd{\mathbf{0.512}}{0.177}
    & \meanstd{\mathbf{24.19}}{4.15}
    & \meanstd{\mathbf{0.539}}{0.192}
    & --
    & \meanstd{\mathbf{0.365}}{0.097}
    & \meanstd{\mathbf{22.80}}{4.36}
    & \meanstd{\mathbf{0.457}}{0.139} \\
    \bottomrule
    \end{tabular}%
    }
\end{table*}

\begin{table*}[t]
    \centering
    \footnotesize
    \setlength{\tabcolsep}{3pt}
    \renewcommand{\arraystretch}{1.05}
    \caption{\textbf{Synthetic off-path detection.}
    Mean $\pm$ sample standard deviation across 25 scene--offset pairs;
    best displayed means are bold.
    $^{*}$ denotes the wrap-aware static ego-Doppler lift.}
    \label{tab:supp-synthetic-detection}
    \ablationfit{%
    \begin{tabular}{@{}lcccccc@{}}
    \toprule
    & \multicolumn{6}{c}{Entire area} \\
    \cmidrule(lr){2-7}
    Method
    & CD [m] $\downarrow$
    & CD-Full $\downarrow$
    & IoU@1\,m $\uparrow$
    & Precision $\uparrow$
    & Recall $\uparrow$
    & F1 $\uparrow$ \\
    \midrule
    RadarSplat$^{*}$
    & \meanstd{5.14}{0.69}
    & \meanstd{0.1773}{0.0351}
    & \meanstd{0.185}{0.093}
    & \meanstd{0.090}{0.051}
    & \meanstd{0.132}{0.091}
    & \meanstd{0.106}{0.065} \\
    RadarFields$^{*}$
    & \meanstd{8.06}{0.86}
    & \meanstd{0.2104}{0.0438}
    & \meanstd{0.099}{0.064}
    & \meanstd{0.022}{0.019}
    & \meanstd{0.028}{0.026}
    & \meanstd{0.024}{0.022} \\
    \methodname-static
    & \meanstd{\mathbf{5.08}}{1.03}
    & \meanstd{0.0885}{0.0237}
    & \meanstd{0.230}{0.095}
    & \meanstd{0.330}{0.131}
    & \meanstd{0.172}{0.093}
    & \meanstd{0.223}{0.109} \\
    \methodname
    & \meanstd{5.13}{1.01}
    & \meanstd{\mathbf{0.0884}}{0.0231}
    & \meanstd{\mathbf{0.236}}{0.098}
    & \meanstd{\mathbf{0.348}}{0.129}
    & \meanstd{\mathbf{0.174}}{0.094}
    & \meanstd{\mathbf{0.227}}{0.109} \\
    \bottomrule
    \end{tabular}%
    }

    \par\vspace{5pt}
    \ablationfit{%
    \begin{tabular}{@{}lccccccc@{}}
    \toprule
    & \multicolumn{3}{c}{Entire area}
    & \multicolumn{4}{c}{Foreground} \\
    \cmidrule(lr){2-4}\cmidrule(lr){5-8}
    Method
    & MMD location $\downarrow$
    & MMD Doppler $\downarrow$
    & MMD power $\downarrow$
    & Hit rate $\uparrow$
    & Miss rate $\downarrow$
    & Density sim. $\uparrow$
    & FP boxes $\downarrow$ \\
    \midrule
    RadarSplat$^{*}$
    & \meanstd{\mathbf{0.2278}}{0.0597}
    & \meanstd{0.6362}{0.3841}
    & \meanstd{5.1296}{1.0148}
    & \meanstd{0.360}{0.238}
    & \meanstd{0.640}{0.238}
    & \meanstd{0.244}{0.153}
    & \meanstd{3.4}{5.0} \\
    RadarFields$^{*}$
    & \meanstd{0.3196}{0.0521}
    & \meanstd{0.7134}{0.5546}
    & \meanstd{5.9267}{1.0330}
    & \meanstd{0.305}{0.199}
    & \meanstd{0.695}{0.199}
    & \meanstd{0.220}{0.111}
    & \meanstd{5.0}{7.4} \\
    \methodname-static
    & \meanstd{0.5109}{0.2132}
    & \meanstd{\mathbf{0.6353}}{0.3296}
    & \meanstd{1.8323}{1.0790}
    & \meanstd{0.224}{0.132}
    & \meanstd{0.776}{0.132}
    & \meanstd{0.240}{0.214}
    & \meanstd{\mathbf{0.7}}{0.8} \\
    \methodname
    & \meanstd{0.5345}{0.2215}
    & \meanstd{0.6398}{0.2961}
    & \meanstd{\mathbf{1.7327}}{1.1115}
    & \meanstd{\mathbf{0.746}}{0.222}
    & \meanstd{\mathbf{0.254}}{0.222}
    & \meanstd{\mathbf{0.341}}{0.155}
    & \meanstd{4.9}{4.5} \\
    \bottomrule
    \end{tabular}%
    }
\end{table*}

\subsection{Runtime}
\label{sec:supp-runtime}

We measure computational cost on RADIal using an otherwise
idle NVIDIA RTX 4090 (Tab.~\ref{tab:supp-runtime}).
\methodname\ requires a median of $7.2$ minutes for 10k
optimization steps and $5.65\,\mathrm{ms}$ to render a full
RAD tensor, with $1055\,\mathrm{MiB}$ peak allocated GPU memory.
Rendering measurements cover the complete forward pass
after warm-up, excluding loading and file output.
The baselines render RA maps, so their latency measures
a different output.
Training times follow each method's stated budget.

\begin{table*}[t]
\centering
\small
\setlength{\tabcolsep}{6pt}
\renewcommand{\arraystretch}{1.05}
\caption{\textbf{Computational cost on RADIal.}
Median training time, rendering latency, and peak allocated
GPU memory on an RTX 4090.}
\label{tab:supp-runtime}
\begin{tabular}{llrrrl}
\toprule
Method
& Training budget
& Train [min]
& Render [ms/frame]
& Peak [MiB]
& Rendered output \\
\midrule
RadarSplat
& 10k steps
& 4.3
& 4.68
& 2138
& RA map \\
RadarFields
& 800 iterations/image
& 1.4
& 15.10
& 4669
& RA map \\
\methodname
& 10k steps
& 7.2
& 5.65
& 1055
& RAD tensor \\
\bottomrule
\end{tabular}
\end{table*}

The two splatting methods have similar rendering latencies,
with \methodname's modest overhead consistent with motion
evaluation and full-RAD synthesis.
RadarFields has higher rendering latency, consistent with
its neural-field rendering approach.

\section{Ablations}
\label{sec:suppl-ablations}
We evaluate the ablations using the RADIal on-path and off-path protocols, applying each ablation in both fitting stages of the off-path cycle.
The learned-extent ablation limits the representation to 300{,}000 primitives.
We report mean $\pm$ sample standard deviation across ten sequences.
Tables~\ref{tab:suppl_abl_onpath_reconstruction}
and~\ref{tab:suppl_abl_onpath_detection} report all on-path ablations;
Tables~\ref{tab:suppl_abl_offpath_reconstruction}
and~\ref{tab:suppl_abl_offpath_detection} report the corresponding
off-path results.

\begin{table*}[t]
    \centering
    \footnotesize
    \setlength{\tabcolsep}{3pt}
    \renewcommand{\arraystretch}{1.05}
    \caption{\textbf{Ablations on-path on RADIal: reconstruction.}
    Mean $\pm$ sample standard deviation across ten sequences;
    best displayed means, including ties, are bold.}
    \label{tab:suppl_abl_onpath_reconstruction}
    \ablationfit{%
    \begin{tabular}{@{}lccccccc@{}}
    \toprule
    & \multicolumn{4}{c}{Full} & \multicolumn{3}{c}{Object} \\
    \cmidrule(lr){2-5}\cmidrule(lr){6-8}
    Method
    & $\rho\uparrow$ & PSNR [dB] $\uparrow$
    & SSIM $\uparrow$ & LPIPS $\downarrow$
    & $\rho\uparrow$ & PSNR [dB] $\uparrow$ & SSIM $\uparrow$ \\
    \midrule
    \multicolumn{8}{l}{\emph{RAD reconstruction}} \\
    w/o Doppler
    & \meanstd{0.226}{0.030}
    & \meanstd{24.889}{0.832}
    & \meanstd{0.448}{0.017}
    & \meanstd{0.283}{0.015}
    & \meanstd{0.504}{0.125}
    & \meanstd{15.374}{1.165}
    & \meanstd{0.292}{0.046} \\
    w/o $\mathcal{L}_{\mathrm{int}}$
    & \meanstd{0.221}{0.089}
    & \meanstd{23.988}{1.510}
    & \meanstd{0.511}{0.024}
    & \meanstd{0.277}{0.019}
    & \meanstd{\mathbf{0.667}}{0.087}
    & \meanstd{17.038}{1.082}
    & \meanstd{\mathbf{0.383}}{0.039} \\
    Learned extent
    & \meanstd{0.263}{0.020}
    & \meanstd{26.114}{0.977}
    & \meanstd{0.535}{0.018}
    & \meanstd{0.567}{0.012}
    & \meanstd{0.610}{0.090}
    & \meanstd{16.801}{1.322}
    & \meanstd{0.360}{0.049} \\
    Learned PSF
    & \meanstd{\mathbf{0.305}}{0.027}
    & \meanstd{\mathbf{26.607}}{0.940}
    & \meanstd{\mathbf{0.547}}{0.012}
    & \meanstd{0.392}{0.017}
    & \meanstd{0.636}{0.089}
    & \meanstd{\mathbf{17.071}}{1.074}
    & \meanstd{0.336}{0.047} \\
    \methodname
    & \meanstd{0.272}{0.062}
    & \meanstd{25.884}{0.951}
    & \meanstd{0.502}{0.022}
    & \meanstd{\mathbf{0.275}}{0.025}
    & \meanstd{0.658}{0.070}
    & \meanstd{16.966}{1.056}
    & \meanstd{0.374}{0.028} \\
    \midrule
    \multicolumn{8}{l}{\emph{RA reconstruction}} \\
    w/o Doppler
    & \meanstd{\mathbf{0.667}}{0.032}
    & \meanstd{\mathbf{26.439}}{1.016}
    & \meanstd{0.682}{0.024}
    & \meanstd{\mathbf{0.217}}{0.016}
    & \meanstd{\mathbf{0.834}}{0.056}
    & \meanstd{\mathbf{15.818}}{0.885}
    & \meanstd{\mathbf{0.616}}{0.040} \\
    w/o $\mathcal{L}_{\mathrm{int}}$
    & \meanstd{0.393}{0.106}
    & \meanstd{21.395}{1.806}
    & \meanstd{0.566}{0.023}
    & \meanstd{0.243}{0.012}
    & \meanstd{0.790}{0.114}
    & \meanstd{14.036}{1.939}
    & \meanstd{0.465}{0.120} \\
    Learned extent
    & \meanstd{0.600}{0.046}
    & \meanstd{25.707}{1.109}
    & \meanstd{\mathbf{0.721}}{0.018}
    & \meanstd{0.477}{0.017}
    & \meanstd{0.783}{0.075}
    & \meanstd{13.415}{0.987}
    & \meanstd{0.494}{0.066} \\
    Learned PSF
    & \meanstd{0.645}{0.039}
    & \meanstd{26.267}{1.059}
    & \meanstd{0.712}{0.020}
    & \meanstd{0.346}{0.018}
    & \meanstd{0.788}{0.081}
    & \meanstd{12.554}{1.184}
    & \meanstd{0.438}{0.055} \\
    \methodname
    & \meanstd{0.612}{0.031}
    & \meanstd{25.780}{1.091}
    & \meanstd{0.690}{0.025}
    & \meanstd{0.234}{0.017}
    & \meanstd{0.827}{0.087}
    & \meanstd{14.927}{1.889}
    & \meanstd{0.568}{0.097} \\
    \midrule
    \multicolumn{8}{l}{\emph{RD reconstruction}} \\
    w/o Doppler
    & \meanstd{0.418}{0.041}
    & \meanstd{24.759}{1.701}
    & \meanstd{0.547}{0.033}
    & --
    & \meanstd{0.440}{0.135}
    & \meanstd{19.854}{2.175}
    & \meanstd{0.489}{0.059} \\
    w/o $\mathcal{L}_{\mathrm{int}}$
    & \meanstd{0.333}{0.135}
    & \meanstd{21.824}{1.721}
    & \meanstd{0.554}{0.041}
    & --
    & \meanstd{0.510}{0.177}
    & \meanstd{19.669}{2.453}
    & \meanstd{0.487}{0.094} \\
    Learned extent
    & \meanstd{0.460}{0.043}
    & \meanstd{25.422}{1.786}
    & \meanstd{\mathbf{0.663}}{0.035}
    & --
    & \meanstd{0.500}{0.111}
    & \meanstd{19.677}{2.273}
    & \meanstd{0.539}{0.104} \\
    Learned PSF
    & \meanstd{\mathbf{0.493}}{0.033}
    & \meanstd{\mathbf{25.717}}{1.770}
    & \meanstd{0.649}{0.032}
    & --
    & \meanstd{0.502}{0.104}
    & \meanstd{19.863}{2.148}
    & \meanstd{0.530}{0.093} \\
    \methodname
    & \meanstd{0.476}{0.067}
    & \meanstd{25.331}{1.700}
    & \meanstd{0.620}{0.040}
    & --
    & \meanstd{\mathbf{0.564}}{0.100}
    & \meanstd{\mathbf{20.786}}{1.948}
    & \meanstd{\mathbf{0.558}}{0.086} \\
    \bottomrule
    \end{tabular}%
    }
\end{table*}
\begin{table*}[t]
    \centering
    \footnotesize
    \setlength{\tabcolsep}{3pt}
    \renewcommand{\arraystretch}{1.05}
    \caption{\textbf{Ablations on-path on RADIal: detection and Doppler.}
    Mean $\pm$ sample standard deviation across ten sequences;
    best displayed means, including ties, are bold.
    Doppler MAE is peak-location error at annotated cars, in bins;
    it is reported here only for the Doppler-supervision comparison.}
    \label{tab:suppl_abl_onpath_detection}
    \ablationfit{%
    \begin{tabular}{@{}lccccccc@{}}
    \toprule
    & \multicolumn{6}{c}{Entire area} & Doppler \\
    \cmidrule(lr){2-7}\cmidrule(lr){8-8}
    Method
    & CD [m] $\downarrow$ & CD-Full $\downarrow$
    & IoU@1\,m $\uparrow$ & Precision $\uparrow$
    & Recall $\uparrow$ & F1 $\uparrow$ & MAE $\downarrow$ \\
    \midrule
    w/o Doppler
    & \meanstd{\mathbf{3.663}}{0.660}
    & \meanstd{\mathbf{0.069}}{0.006}
    & \meanstd{\mathbf{0.240}}{0.049}
    & \meanstd{0.142}{0.025}
    & \meanstd{\mathbf{0.153}}{0.031}
    & \meanstd{0.144}{0.021}
    & \meanstd{1.843}{0.585} \\
    w/o $\mathcal{L}_{\mathrm{int}}$
    & \meanstd{3.974}{0.877}
    & \meanstd{0.070}{0.010}
    & \meanstd{0.193}{0.057}
    & \meanstd{0.162}{0.065}
    & \meanstd{0.152}{0.058}
    & \meanstd{0.155}{0.061}
    & -- \\
    Learned extent
    & \meanstd{7.468}{2.520}
    & \meanstd{0.100}{0.015}
    & \meanstd{0.126}{0.039}
    & \meanstd{0.224}{0.052}
    & \meanstd{0.067}{0.024}
    & \meanstd{0.101}{0.032}
    & -- \\
    Learned PSF
    & \meanstd{6.302}{1.762}
    & \meanstd{0.114}{0.025}
    & \meanstd{0.112}{0.032}
    & \meanstd{\mathbf{0.243}}{0.031}
    & \meanstd{0.058}{0.022}
    & \meanstd{0.089}{0.028}
    & -- \\
    \methodname
    & \meanstd{4.008}{1.021}
    & \meanstd{\mathbf{0.069}}{0.009}
    & \meanstd{0.238}{0.040}
    & \meanstd{0.233}{0.051}
    & \meanstd{0.149}{0.037}
    & \meanstd{\mathbf{0.179}}{0.041}
    & \meanstd{\mathbf{0.732}}{0.290} \\
    \bottomrule
    \end{tabular}%
    }

    \par\vspace{5pt}
    \ablationfit{%
    \begin{tabular}{@{}lccccccc@{}}
    \toprule
    & \multicolumn{3}{c}{Entire area}
    & \multicolumn{4}{c}{Foreground} \\
    \cmidrule(lr){2-4}\cmidrule(lr){5-8}
    Method
    & MMD location $\downarrow$ & MMD Doppler $\downarrow$
    & MMD power $\downarrow$ & Hit rate $\uparrow$
    & Miss rate $\downarrow$ & Density sim. $\uparrow$
    & FP boxes $\downarrow$ \\
    \midrule
    w/o Doppler
    & \meanstd{0.541}{0.254}
    & \meanstd{0.471}{0.166}
    & \meanstd{0.834}{0.301}
    & \meanstd{0.909}{0.099}
    & \meanstd{0.091}{0.099}
    & \meanstd{0.530}{0.107}
    & \meanstd{\mathbf{0.200}}{0.422} \\
    w/o $\mathcal{L}_{\mathrm{int}}$
    & \meanstd{0.496}{0.254}
    & \meanstd{\mathbf{0.095}}{0.050}
    & \meanstd{0.582}{0.197}
    & \meanstd{\mathbf{0.910}}{0.119}
    & \meanstd{\mathbf{0.090}}{0.119}
    & \meanstd{0.547}{0.113}
    & \meanstd{\mathbf{0.200}}{0.422} \\
    Learned extent
    & \meanstd{0.545}{0.223}
    & \meanstd{0.133}{0.049}
    & \meanstd{\mathbf{0.405}}{0.132}
    & \meanstd{0.842}{0.130}
    & \meanstd{0.158}{0.130}
    & \meanstd{0.518}{0.130}
    & \meanstd{\mathbf{0.200}}{0.422} \\
    Learned PSF
    & \meanstd{0.694}{0.361}
    & \meanstd{0.529}{0.215}
    & \meanstd{1.001}{0.351}
    & \meanstd{0.891}{0.118}
    & \meanstd{0.109}{0.118}
    & \meanstd{0.396}{0.102}
    & \meanstd{\mathbf{0.200}}{0.422} \\
    \methodname
    & \meanstd{\mathbf{0.421}}{0.142}
    & \meanstd{0.166}{0.074}
    & \meanstd{0.593}{0.208}
    & \meanstd{0.907}{0.086}
    & \meanstd{0.093}{0.086}
    & \meanstd{\mathbf{0.583}}{0.099}
    & \meanstd{0.300}{0.483} \\
    \bottomrule
    \end{tabular}%
    }
\end{table*}

\begin{table*}[t]
    \centering
    \footnotesize
    \setlength{\tabcolsep}{3pt}
    \renewcommand{\arraystretch}{1.05}
    \caption{\textbf{Ablations off-path on RADIal: reconstruction.}
    Results after the $2\,\mathrm{m}$ lateral-shift cycle.
    Mean $\pm$ sample standard deviation across ten sequences;
    best displayed means, including ties, are bold.}
    \label{tab:suppl_abl_offpath_reconstruction}
    \ablationfit{%
    \begin{tabular}{@{}lccccccc@{}}
    \toprule
    & \multicolumn{4}{c}{Full} & \multicolumn{3}{c}{Object} \\
    \cmidrule(lr){2-5}\cmidrule(lr){6-8}
    Method
    & $\rho\uparrow$ & PSNR [dB] $\uparrow$
    & SSIM $\uparrow$ & LPIPS $\downarrow$
    & $\rho\uparrow$ & PSNR [dB] $\uparrow$ & SSIM $\uparrow$ \\
    \midrule
    \multicolumn{8}{l}{\emph{RAD reconstruction}} \\
    w/o Doppler
    & \meanstd{0.208}{0.020}
    & \meanstd{24.676}{0.745}
    & \meanstd{0.419}{0.015}
    & \meanstd{0.295}{0.012}
    & \meanstd{0.389}{0.074}
    & \meanstd{14.610}{1.271}
    & \meanstd{0.264}{0.034} \\
    w/o $\mathcal{L}_{\mathrm{int}}$
    & \meanstd{\mathbf{0.447}}{0.062}
    & \meanstd{\mathbf{27.153}}{0.913}
    & \meanstd{0.548}{0.016}
    & \meanstd{\mathbf{0.252}}{0.017}
    & \meanstd{\mathbf{0.549}}{0.046}
    & \meanstd{15.890}{1.401}
    & \meanstd{\mathbf{0.356}}{0.053} \\
    Learned extent
    & \meanstd{0.313}{0.031}
    & \meanstd{26.712}{0.880}
    & \meanstd{\mathbf{0.551}}{0.020}
    & \meanstd{0.609}{0.013}
    & \meanstd{0.424}{0.080}
    & \meanstd{15.574}{1.204}
    & \meanstd{0.307}{0.045} \\
    Learned PSF
    & \meanstd{0.349}{0.026}
    & \meanstd{26.938}{0.875}
    & \meanstd{0.539}{0.011}
    & \meanstd{0.381}{0.010}
    & \meanstd{0.488}{0.116}
    & \meanstd{15.849}{1.213}
    & \meanstd{0.318}{0.049} \\
    \methodname
    & \meanstd{0.363}{0.052}
    & \meanstd{26.534}{0.736}
    & \meanstd{0.498}{0.012}
    & \meanstd{0.267}{0.015}
    & \meanstd{0.536}{0.077}
    & \meanstd{\mathbf{15.977}}{1.357}
    & \meanstd{0.344}{0.041} \\
    \midrule
    \multicolumn{8}{l}{\emph{RA reconstruction}} \\
    w/o Doppler
    & \meanstd{\mathbf{0.694}}{0.041}
    & \meanstd{\mathbf{26.762}}{1.210}
    & \meanstd{0.675}{0.021}
    & \meanstd{0.232}{0.018}
    & \meanstd{\mathbf{0.835}}{0.040}
    & \meanstd{\mathbf{15.567}}{0.778}
    & \meanstd{\mathbf{0.595}}{0.042} \\
    w/o $\mathcal{L}_{\mathrm{int}}$
    & \meanstd{0.655}{0.054}
    & \meanstd{24.655}{0.928}
    & \meanstd{0.556}{0.023}
    & \meanstd{\mathbf{0.203}}{0.014}
    & \meanstd{0.818}{0.049}
    & \meanstd{13.590}{1.775}
    & \meanstd{0.438}{0.083} \\
    Learned extent
    & \meanstd{0.640}{0.024}
    & \meanstd{26.190}{1.056}
    & \meanstd{\mathbf{0.729}}{0.017}
    & \meanstd{0.532}{0.022}
    & \meanstd{0.732}{0.083}
    & \meanstd{12.712}{1.390}
    & \meanstd{0.413}{0.082} \\
    Learned PSF
    & \meanstd{0.665}{0.022}
    & \meanstd{26.419}{1.043}
    & \meanstd{0.698}{0.020}
    & \meanstd{0.344}{0.016}
    & \meanstd{0.761}{0.075}
    & \meanstd{12.581}{1.506}
    & \meanstd{0.409}{0.079} \\
    \methodname
    & \meanstd{0.661}{0.035}
    & \meanstd{26.150}{0.951}
    & \meanstd{0.665}{0.023}
    & \meanstd{0.212}{0.015}
    & \meanstd{0.807}{0.057}
    & \meanstd{14.138}{1.662}
    & \meanstd{0.517}{0.089} \\
    \midrule
    \multicolumn{8}{l}{\emph{RD reconstruction}} \\
    w/o Doppler
    & \meanstd{0.388}{0.037}
    & \meanstd{24.577}{1.485}
    & \meanstd{0.519}{0.030}
    & --
    & \meanstd{0.305}{0.082}
    & \meanstd{18.988}{2.233}
    & \meanstd{0.415}{0.069} \\
    w/o $\mathcal{L}_{\mathrm{int}}$
    & \meanstd{\mathbf{0.746}}{0.029}
    & \meanstd{25.604}{1.008}
    & \meanstd{0.596}{0.032}
    & --
    & \meanstd{\mathbf{0.679}}{0.073}
    & \meanstd{20.264}{1.756}
    & \meanstd{0.526}{0.051} \\
    Learned extent
    & \meanstd{0.550}{0.034}
    & \meanstd{26.010}{1.690}
    & \meanstd{\mathbf{0.689}}{0.032}
    & --
    & \meanstd{0.523}{0.114}
    & \meanstd{19.682}{2.194}
    & \meanstd{\mathbf{0.551}}{0.092} \\
    Learned PSF
    & \meanstd{0.585}{0.024}
    & \meanstd{26.203}{1.705}
    & \meanstd{0.671}{0.029}
    & --
    & \meanstd{0.545}{0.099}
    & \meanstd{19.936}{2.174}
    & \meanstd{0.537}{0.087} \\
    \methodname
    & \meanstd{0.653}{0.048}
    & \meanstd{\mathbf{26.616}}{1.256}
    & \meanstd{0.661}{0.022}
    & --
    & \meanstd{0.584}{0.094}
    & \meanstd{\mathbf{20.561}}{1.973}
    & \meanstd{\mathbf{0.551}}{0.053} \\
    \bottomrule
    \end{tabular}%
    }
\end{table*}

\begin{table*}[t]
    \centering
    \footnotesize
    \setlength{\tabcolsep}{3pt}
    \renewcommand{\arraystretch}{1.05}
    \caption{\textbf{Ablations off-path on RADIal: detection and Doppler.}
    Results after the $2\,\mathrm{m}$ lateral-shift cycle.
    Mean $\pm$ sample standard deviation across ten sequences;
    best displayed means, including ties, are bold.
    Doppler MAE is peak-location error at annotated cars, in bins;
    it is reported here only for the Doppler-supervision comparison.}
    \label{tab:suppl_abl_offpath_detection}
    \ablationfit{%
    \begin{tabular}{@{}lccccccc@{}}
    \toprule
    & \multicolumn{6}{c}{Entire area} & Doppler \\
    \cmidrule(lr){2-7}\cmidrule(lr){8-8}
    Method
    & CD [m] $\downarrow$ & CD-Full $\downarrow$
    & IoU@1\,m $\uparrow$ & Precision $\uparrow$
    & Recall $\uparrow$ & F1 $\uparrow$ & MAE $\downarrow$ \\
    \midrule
    w/o Doppler
    & \meanstd{3.586}{0.447}
    & \meanstd{0.068}{0.005}
    & \meanstd{0.210}{0.045}
    & \meanstd{0.109}{0.021}
    & \meanstd{0.139}{0.023}
    & \meanstd{0.120}{0.018}
    & \meanstd{2.264}{0.511} \\
    w/o $\mathcal{L}_{\mathrm{int}}$
    & \meanstd{\mathbf{2.676}}{0.565}
    & \meanstd{\mathbf{0.063}}{0.010}
    & \meanstd{\mathbf{0.335}}{0.057}
    & \meanstd{\mathbf{0.292}}{0.055}
    & \meanstd{\mathbf{0.305}}{0.062}
    & \meanstd{\mathbf{0.296}}{0.057}
    & -- \\
    Learned extent
    & \meanstd{9.419}{3.427}
    & \meanstd{0.113}{0.020}
    & \meanstd{0.091}{0.042}
    & \meanstd{0.192}{0.077}
    & \meanstd{0.047}{0.023}
    & \meanstd{0.074}{0.034}
    & -- \\
    Learned PSF
    & \meanstd{5.220}{1.232}
    & \meanstd{0.094}{0.016}
    & \meanstd{0.142}{0.031}
    & \meanstd{0.241}{0.051}
    & \meanstd{0.087}{0.023}
    & \meanstd{0.124}{0.031}
    & -- \\
    \methodname
    & \meanstd{3.246}{0.569}
    & \meanstd{\mathbf{0.063}}{0.008}
    & \meanstd{0.268}{0.034}
    & \meanstd{0.251}{0.049}
    & \meanstd{0.227}{0.034}
    & \meanstd{0.236}{0.040}
    & \meanstd{\mathbf{1.375}}{0.431} \\
    \bottomrule
    \end{tabular}%
    }

    \par\vspace{5pt}
    \ablationfit{%
    \begin{tabular}{@{}lccccccc@{}}
    \toprule
    & \multicolumn{3}{c}{Entire area}
    & \multicolumn{4}{c}{Foreground} \\
    \cmidrule(lr){2-4}\cmidrule(lr){5-8}
    Method
    & MMD location $\downarrow$ & MMD Doppler $\downarrow$
    & MMD power $\downarrow$ & Hit rate $\uparrow$
    & Miss rate $\downarrow$ & Density sim. $\uparrow$
    & FP boxes $\downarrow$ \\
    \midrule
    w/o Doppler
    & \meanstd{0.496}{0.266}
    & \meanstd{0.436}{0.170}
    & \meanstd{1.027}{0.299}
    & \meanstd{0.927}{0.071}
    & \meanstd{0.073}{0.071}
    & \meanstd{\mathbf{0.532}}{0.105}
    & \meanstd{1.000}{1.247} \\
    w/o $\mathcal{L}_{\mathrm{int}}$
    & \meanstd{\mathbf{0.295}}{0.222}
    & \meanstd{\mathbf{0.080}}{0.035}
    & \meanstd{1.588}{0.376}
    & \meanstd{\mathbf{0.944}}{0.083}
    & \meanstd{\mathbf{0.056}}{0.083}
    & \meanstd{0.450}{0.118}
    & \meanstd{1.700}{1.889} \\
    Learned extent
    & \meanstd{0.794}{0.349}
    & \meanstd{0.117}{0.021}
    & \meanstd{\mathbf{0.370}}{0.082}
    & \meanstd{0.587}{0.255}
    & \meanstd{0.413}{0.255}
    & \meanstd{0.328}{0.143}
    & \meanstd{\mathbf{0.200}}{0.422} \\
    Learned PSF
    & \meanstd{0.561}{0.247}
    & \meanstd{0.341}{0.150}
    & \meanstd{1.287}{0.341}
    & \meanstd{0.776}{0.234}
    & \meanstd{0.224}{0.234}
    & \meanstd{0.368}{0.115}
    & \meanstd{0.800}{0.789} \\
    \methodname
    & \meanstd{0.353}{0.161}
    & \meanstd{0.175}{0.082}
    & \meanstd{1.203}{0.211}
    & \meanstd{0.917}{0.101}
    & \meanstd{0.083}{0.101}
    & \meanstd{0.521}{0.111}
    & \meanstd{1.100}{1.197} \\
    \bottomrule
    \end{tabular}%
    }
\end{table*}

\section{Sensor Configuration Transfer}
\label{sec:supp-config-transfer}

We evaluate coarse-to-fine sensor-configuration transfer against linear
upsampling of the same model's coarse renders.
Both measurement configurations are derived from the same RADIal raw
ADC recordings; the scene and acquisition platform remain unchanged.

\paragraph{Configurations.}
The native (fine) configuration uses 512 ADC samples per chirp and
256 chirps per coherent processing interval.
The coarse configuration retains 128 samples per chirp and
128 consecutive chirps, yielding fourfold coarser range resolution and
twofold coarser Doppler resolution.
The native and coarse RAD tensors contain $16\times447\times751$ and $8\times112\times751$ bins after range cropping, respectively.
The azimuth grid and Doppler wrap period of $1.7968\,\mathrm{m/s}$ remain unchanged.
Both sets of measurements are processed directly from raw ADC;
the coarse measurements are not obtained by resizing native tensors.

\paragraph{Training and rendering.}
We fit a model to the coarse measurements for each of the ten sequences
in the RADIal on-path split, using 10k optimization steps.
Initialization uses only the coarse training measurements, with
configuration-specific normalization.
For direct rendering, we replace the coarse PSF and sampling grid
with those of the native configuration, keeping all reflector and
motion parameters fixed.
For linear upsampling, we render the same model under the coarse
configuration and interpolate along range and Doppler, using circular
interpolation over the shared Doppler wrap period.

\begin{figure*}[t]
    \centering
    \includegraphics[width=0.75\textwidth]
        {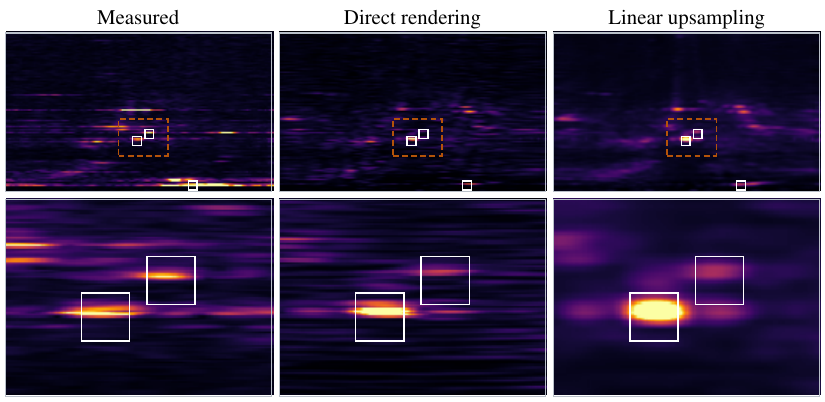}
    \caption{\textbf{Changing the measurement configuration
    of a reconstructed scene.}
    Left: measured native RA projection.
    Middle: the coarse-trained scene rendered with the native PSF
    and sampling grid.
    Right: linear upsampling of the same scene's coarse render.
    The bottom row enlarges the dashed region; white boxes mark vehicles.
}
    \label{fig:supp-config-transfer}
\end{figure*}

\paragraph{Evaluation.}
Both outputs are evaluated against the same real held-out native
measurements using the same scorer.
Table~\ref{tab:supp-config-transfer} reports reconstruction and detection
results as means and sample standard deviations across the ten sequences.
Figure~\ref{fig:supp-config-transfer} compares the two rendering routes
with the native measurements.

\begin{table*}[t]
    \centering
    \footnotesize
    \setlength{\tabcolsep}{3pt}
    \renewcommand{\arraystretch}{1.05}
    \caption{\textbf{Coarse-to-fine configuration transfer
    versus linear upsampling.}
    Both outputs use the same coarse-trained model and are
    evaluated against held-out native measurements.
    Mean $\pm$ sample standard deviation across ten RADIal sequences.
    Object metrics use annotated vehicle regions.
    Best displayed means are bold.
    LPIPS is not reported for RD.}
    \label{tab:supp-config-transfer}

    \ablationfit{%
    \begin{tabular}{@{}lcccccc@{}}
        \toprule
        & \multicolumn{3}{c}{Full}
        & \multicolumn{3}{c}{Object} \\
        \cmidrule(lr){2-4}\cmidrule(lr){5-7}
        Method
        & $\rho\uparrow$
        & SSIM $\uparrow$
        & LPIPS $\downarrow$
        & $\rho\uparrow$
        & PSNR [dB] $\uparrow$
        & SSIM $\uparrow$ \\
        \midrule
        \multicolumn{7}{l}{\emph{RAD reconstruction}} \\
        Linear upsampling
        & \meanstd{\mathbf{0.329}}{0.021}
        & \meanstd{\mathbf{0.595}}{0.014}
        & \meanstd{0.540}{0.008}
        & \meanstd{\mathbf{0.556}}{0.044}
        & \meanstd{12.87}{2.70}
        & \meanstd{\mathbf{0.315}}{0.058} \\
        Direct rendering
        & \meanstd{0.201}{0.037}
        & \meanstd{0.461}{0.012}
        & \meanstd{\mathbf{0.263}}{0.014}
        & \meanstd{0.522}{0.103}
        & \meanstd{\mathbf{15.32}}{1.10}
        & \meanstd{0.301}{0.040} \\
        \midrule
        \multicolumn{7}{l}{\emph{RA reconstruction}} \\
        Linear upsampling
        & \meanstd{\mathbf{0.624}}{0.033}
        & \meanstd{\mathbf{0.726}}{0.019}
        & \meanstd{0.472}{0.012}
        & \meanstd{\mathbf{0.629}}{0.056}
        & \meanstd{10.30}{2.08}
        & \meanstd{\mathbf{0.297}}{0.084} \\
        Direct rendering
        & \meanstd{0.501}{0.035}
        & \meanstd{0.549}{0.015}
        & \meanstd{\mathbf{0.166}}{0.011}
        & \meanstd{0.528}{0.144}
        & \meanstd{\mathbf{11.15}}{1.03}
        & \meanstd{0.285}{0.070} \\
        \midrule
        \multicolumn{7}{l}{\emph{RD reconstruction}} \\
        Linear upsampling
        & \meanstd{\mathbf{0.516}}{0.046}
        & \meanstd{\mathbf{0.666}}{0.040}
        & --
        & \meanstd{\mathbf{0.513}}{0.058}
        & \meanstd{\mathbf{19.82}}{2.35}
        & \meanstd{\mathbf{0.536}}{0.097} \\
        Direct rendering
        & \meanstd{0.361}{0.055}
        & \meanstd{0.514}{0.032}
        & --
        & \meanstd{0.380}{0.092}
        & \meanstd{18.87}{2.01}
        & \meanstd{0.431}{0.075} \\
        \bottomrule
    \end{tabular}%
    }

    \par\vspace{5pt}

    \ablationfit{%
    \begin{tabular}{@{}lccccc@{}}
        \toprule
        \multicolumn{6}{l}{\emph{Detection}} \\
        Method
        & CD [m] $\downarrow$
        & IoU@1\,m $\uparrow$
        & F1 $\uparrow$
        & Hit rate $\uparrow$
        & Density sim. $\uparrow$ \\
        \midrule
        Linear upsampling
        & \meanstd{7.27}{1.71}
        & \meanstd{0.103}{0.035}
        & \meanstd{0.101}{0.038}
        & \meanstd{0.906}{0.124}
        & \meanstd{0.339}{0.077} \\
        Direct rendering
        & \meanstd{\mathbf{3.20}}{0.54}
        & \meanstd{\mathbf{0.281}}{0.035}
        & \meanstd{\mathbf{0.195}}{0.046}
        & \meanstd{\mathbf{0.960}}{0.074}
        & \meanstd{\mathbf{0.593}}{0.076} \\
        \bottomrule
    \end{tabular}%
    }
\end{table*}

\end{document}